\documentclass[manuscript]{acmart}

\usepackage{bbding}

\usepackage{makecell,multirow}

\usepackage{wrapfig}

\AtBeginDocument{%
  }

\setcopyright{acmlicensed}
\copyrightyear{2025}
\acmYear{2025}
\acmDOI{XXXXXXX.XXXXXXX}

\acmJournal{CSUR}

\begin{document}

%\title[Interpretability of linguistic knowledge inside Pre-trained Transformer-based Models: a literature review]{Interpretability of linguistic knowledge inside Pre-trained Transformer-based Language Models: a systematic literature review}

\title[Linguistic Interpretability of Transformer-based Language Models]{Linguistic Interpretability of Transformer-based Language Models: a systematic review}

\author{Miguel López-Otal}
\email{mlopezotal@unizar.es}
\orcid{0000-0002-4854-8062}
\affiliation{
  \institution{Aragon Institute of Engineering Research, University of Zaragoza}
  \city{Zaragoza}
  \country{Spain}
}

\author{Jorge Gracia}
\email{jogracia@unizar.es}
\orcid{0000-0001-6452-7627}
\affiliation{
  \institution{Aragon Institute of Engineering Research, University of Zaragoza}
  \city{Zaragoza}
  \country{Spain}
  }

\author{Jordi Bernad}
\email{jbernad@unizar.es}
\orcid{0000-0001-8531-353X}
\affiliation{
  \institution{Aragon Institute of Engineering Research, University of Zaragoza}
  \city{Zaragoza}
  \country{Spain}
}

\author{Carlos Bobed}
\email{cbobed@unizar.es}
\orcid{0000-0003-4239-8785}
\affiliation{
 \institution{Aragon Institute of Engineering Research, University of Zaragoza}
  \city{Zaragoza}
  \country{Spain}}

\author{Lucía Pitarch-Ballesteros}
\email{lpitarch@unizar.es}
\orcid{0000-0002-6734-8808}
\affiliation{
  \institution{Aragon Institute of Engineering Research, University of Zaragoza}
  \city{Zaragoza}
  \country{Spain}}

\author{Emma Anglés-Herrero}
\email{emmaa.herrero@gmail.com}
\orcid{0009-0000-6403-2202}
\affiliation{
  \institution{Aragon Institute of Engineering Research, University of Zaragoza}
  \city{Zaragoza}
  \country{Spain}}

\renewcommand{\shortauthors}{López-Otal et al.}

\begin{abstract}
Language models based on the Transformer architecture achieve excellent results in many language-related tasks, such as text classification or sentiment analysis. However, despite the architecture of these models being well-defined, little is known about how their internal computations help them achieve their results. This renders these models, as of today, a type of `black box' systems. There is, however, a line of research --`interpretability'-- aiming to learn how information is encoded inside these models. More specifically, there is work dedicated to studying whether Transformer-based models possess knowledge of linguistic phenomena similar to human speakers –-an area we call `linguistic interpretability' of these models.

In this survey we present a comprehensive analysis of 160 research works, spread across multiple languages and models --including multilingual ones--, that attempt to discover linguistic information from the perspective of several traditional Linguistics disciplines: Syntax, Morphology, Lexico-Semantics and Discourse. Our survey fills a gap in the existing interpretability literature, which either not focus on linguistic knowledge in these models or present some limitations --e.g. only studying English-based models. Our survey also focuses on Pre-trained Language Models not further specialized for a downstream task, with an emphasis on works that use interpretability techniques that explore models' internal representations.

\end{abstract}





\maketitle

\section{Introduction}

Language models (LMs) based on the Transformer architecture~\citep{Vaswani}, have become the state-of-the-art for many downstream natural language processing (NLP) tasks, such as text classification, sentiment analysis or summarization. Outside of the research community and industry use, the general public has become increasingly aware of this technology's potential thanks to popular chatbot tools, such as ChatGPT~\citep{OpenAI}. 
Along a wide adoption of these models, there has also been a growing interest in understanding their real capabilities as well as their limitations. Although the architecture of Transformer-based models is well-defined, when a model of this type is deployed, we are unsure what sort of internal operations happen within that lead them to achieve such accuracy in many tasks across many domains.\footnote{In this sense, it is not too dissimilar from the study of the human brain: while we know how an individual synapse fires across neurons, we do not know for certain how this can eventually lead to more complex notions such as structured reasoning or the emergence of abstract concepts in human thinkers.}

The implementation of Transformer-based models is usually based first on their pre-training on large corpora of text –-from which they apparently acquire most of their knowledge to accomplish their goals–- and then on its specialization on an end task –-e.g. fine-tuning. These systems are pre-trained in a self-supervised way, with no explicit control by humans over how they should modify their internal representations to achieve their end result, beyond training metrics selection and the setting of hyperparameters. 

This renders the Transformer-based architecture a type of ‘black box’: it performs a series of tasks without the user truly being aware on how their internal implementation allows them to do so. This becomes an issue as modern language models may have implicitly learned social biases or incorrect facts from texts seen during pre-training, a source of information that can be non-trivial to locate or neutralize. As such, understanding and explaining these models becomes ``crucial for elucidating their behaviors, limitations, and social impacts''~\citep{Explainability_for_Large_Language_Models_A_Survey}.

In this sense, a broad line of research has been interested in the understanding of the internal operations happening inside Transformer models, as well as other machine learning architectures. One of those lines, `explainability', attempts to \textit{explain} --in human terms-- how a model has accomplished a specific task, in order to gain knowledge on their potential biases~\citep{XAIvsIAI}. Another related area --in which we are more interested in the context of this work-- is that of `interpretability', which also aims to grasp an understanding on how a model achieves a result but, more specifically, it attempts at the same time to \textit{interpret} the mechanisms happening inside a model, step-by-step, that exactly lead them towards a specific output or another~\citep{XAIvsIAI, Explainability_for_Large_Language_Models_A_Survey,XAI2,XAI3,XAI4}.\footnote{At times the literature may use the terms `explainability' and `interpretability' in an interchangeable way.} Both explainability and interpretability works might be interested, for instance, in discovering the amount of encyclopedic facts that these models possess or whether they are sensible to reasoning abilities involving commonsense knowledge. The relevance of interpretability works is that they represent attempts to `open up' the so-called \textit{black box} of Transformer-based models, in order to devise in a detailed and technical way how they have come to achieve their state-of-the-art downstream results across many tasks~\citep{XAIvsIAI}.

Within the field of general knowledge interpretability in Transformer Language Models, there is a growing interest in discovering whether these models have been able to acquire a generalization capability of language similar to that of humans. These research works attempt to discern whether these models, through the combination of their self-attention architecture and their exposure to large amounts of text, are able to acquire an internal knowledge of linguistic structures and phenomena (syntax, morphology, semantics, etc.) similar to what is observed in human speakers. This notion is derived from the observation of models accomplishing excellent results in language-related tasks. 
For the purpose of this work, we call this area of studies `\textit{linguistic} interpretability' of these models.

This is a research area that was sparked by work such as the one by~\citet{39e608673194e619191dbda4a5f348a8af0dca9b6603e901c28598351c10afb5}, which was able to uncover partial syntactic dependencies from pairs of contextual embeddings in BERT~\citep{BERT} –-a popular Transformer language model–-, correctly deducing from them the depth of each word in a parse tree as well as the distance between each pair of words. This work demonstrated that syntactic structure may have been indirectly encoded in Transformer-based models simply from their exposure to raw text during pre-training. This led to a number of studies that have attempted to discover whether linguistic information of other types (morphological, semantic, etc.) may have been acquired as well by these models during their training.

This is a controversial topic, as many works~\citep{bender-koller-2020-climbing, stochastic_parrots} claim that Transformer-based models may simply be learning statistical correlations between co-occurring individual tokens, without arriving at a true generalization of any linguistic relationship between them, likening these models to `stochastic parrots'~\citep{stochastic_parrots}. Regardless of this view, some research papers have attempted to analyze the models' internal parameters to discover a sort of generalization capability that can be approximated --although not exactly matched-- to humans' knowledge of language. 

The motivation for uncovering a supposed linguistic competence in modern language models is varied. Other than the knowledge that it provides on the internal implementations of these models, it also brings some clear practical advantages. For instance, while it is only feasible to train a Transformer-based model for a specific language if there exists an immensely large text corpora available for it~\citep{billionsofwordsofpretraining} --which becomes a barrier for low-resourced languages--, having a knowledge on how existing models may process linguistic information on their respective languages could become a source of vital data for attempting to rework one of those models --via various techniques, such as knowledge injection~\citep{know-injection, madX}-- to support a minority language. Also, by providing an insight into some aspects of the pre-training of these models, we might also be able to understand and even manually control some parts of it, something that could help alleviate the environmental and budgeting issues that are commonly associated with the pre-training from scratch of ever-larger models~\citep{stochastic_parrots}. It may also help detect issues with existing machine translation systems based on the Transformer architecture. Additionally, the acquisition of linguistic knowledge in Transformer-based models may give some insights into how humans process and acquire language themselves, thus contributing to a combined study area in linguistic competence in humans and machines --e.g.~\citep{twentyfour_04}.

The information on linguistic competence in Transformer language models, while potentially useful and interesting, is unfortunately sparse and spread across many research works in the literature. As such, recovering information on the topic can become difficult for the researcher, since the general tendency is for an individual paper to either simply report on a single linguistic phenomenon –-e.g. subject-verb agreement–- or, even when discussing many linguistic phenomena, only provide results on a limited number of languages –-mostly English-- or on very few Transformer-based models. This leads to a plethora of studies that report individual instances of language discovery in these models, but without any sort of unifying vision on the overall linguistic competence of the Transformer architecture across languages.

In this survey we aim to provide a unified vision of the conclusions reached by a large body of work aiming to discuss the topic of `linguistic interpretability' --i.e. discovery of linguistic knowledge-- in Transformer-based pre-trained language models, also known as PLMs, or large language models (LLMs) when they are of significant size. We analyze a series of research papers on the topic, amounting to a total of 160 works, across different architectures and typologically-different languages, in order to give a general conclusion on how linguistic information may be present and processed internally within these models. We also present the methodologies commonly used by these works to discover this kind of knowledge, as well as the type of linguistic knowledge and phenomena that are usually investigated. 
The papers and techniques discussed in the scope of this survey complement other existing surveys on PLMs' interpretability~\citep{Explainability_for_Large_Language_Models_A_Survey, XAILING2024, bertology} (see Section~\ref{related_work}), which are more generic and do not put the focus on the linguistic capabilities of these models --or, in the case they do, it is only done partially~\citep{bertology} or relying on interpretability methodologies different to the ones we use in our case~\citep{XAILING2024}.

Our focus is exclusively on those research works that perform a study of Transformer-based models and, most specifically, do it from within; that is, those that analyze for this purpose the internal parameters and intermediate representations of these models, e.g., the embeddings extracted from different layers in a model or attention heads. We do not consider studies that analyze the linguistic performance of a model in a `black box' setting, such as only examining the output of a model's final layer or reporting on the results of a linguistically-motivated classification task. In this sense, we are interested in the so-called \textit{layerwise} performance of these models.

We also only present studies that leverage base, pre-trained models as their object of study, keeping their internal parameters unchanged. We omit studies on models that are fine-tuned --either for a downstream task or even for a linguistic knowledge discovery task--, or on models that have architectural modifications or are otherwise specialized for downstream tasks --such as sentiment analysis or Natural Language Inference (NLI). In that way, we will be able to assess the fundamental or ``baseline'' capacity of pre-trained models to acquire linguistic knowledge, without the interference of additional tasks processing that might introduce some biases in the analysis. Additionally, we do not cover the study of language models that have been adapted for their use as conversational agents via prompts --e.g. ChatGPT--, since those constitute modified models akin to fine-tuned ones. 

Overall, we are interested in the study of linguistic knowledge in base, unmodified pre-trained models, prior to any sort of adaptation to an end task, and on research work that attempts to perform this analysis with methodologies that work from within the models themselves.

The contributions of our work are as follows:
\begin{itemize}

    \item We identify a body of recent and ongoing work (160 papers) on \textit{`linguistic interpretability' within Transformer-based PLMs} focused on techniques that explore the models' internal representations.  

     \item We present such existing research on `linguistic interpretability', which is generally sparse and reporting on miscellaneous linguistic phenomena and languages, under a unified vision, aiming to provide a generalized view of linguistic competence in PLMs across multiple languages.
        
    \item We quantitatively and qualitatively analyze our obtained results along several linguistic levels: semantics, syntax, morphology and discourse.
    
\end{itemize}

The rest of our paper is organized as follows. In Section~\ref{systematic-review-methodology}, we present the systematic review method used in this survey. We provide a small overview of the Transformer architecture in Section~\ref{transformer-elements-definition}. Then, in Section~\ref{quantitative-analysis} we perform a quantitative analysis of our retrieved body of work. In Section~\ref{analysis_ling_competence}, we present our obtained conclusions on the overall linguistic capacities of Transformer-based PLMs. In Section~\ref{discussion} we provide a discussion of the results and in Section~\ref{related_work} we refer to similar works. Concluding remarks and future lines of work can be found in Section~\ref{conclusions}.

\section{Systematic review methodology}
\label{systematic-review-methodology}

Owing to the popularity of Large Language Models, there is an equally large number of research papers that deal into their potential interpretability. 
In order to obtain an overview of this vast field of research, we adopted a clear and straightforward methodology for the retrieval and classification of research papers that were of our interest. This strategy was based on the PRISMA methodology~\citep{PRISMA}, an approach to systematic literature reviews that presents a series of recommended steps that researchers can follow in order to facilitate their task of retrieving and reporting relevant research work for their area of interest. The PRISMA methodology consists of three major steps: identification, screening, and inclusion. These were followed by the detailed analysis of the selected papers.

\subsection{Identification}

We first set out to locate as many research papers as possible on the topic, relying on the following complex keyword-based search, which was reached upon several iterations and trial and errors\footnote{This final list of keywords was decided based on some preliminary experiments with a series of different search queries that were made and measured against a small, hand-crafted corpus of selected articles of the type that we wished to retrieve.}:

\begin{quote}
    transformer OR BERT AND probe* OR interpret* OR explain* OR explan* OR ``psycholinguistic'' OR ``linguistic knowledge'' OR ``linguistic information'' OR ``linguistic property''  AND syntax* OR semantic* OR lexical* OR morphology* OR morphosyntactic* OR grammatical*
\end{quote}

We used Google Scholar\footnote{\url{https://scholar.google.com/}} for retrieving the initial list of papers. We limited our search to articles published between 2018 
and October 2024. This timeframe covers the initial steps of the Transformer-based LMs boom and the subsequent development and creation of newer, ever-growing and larger pre-trained models. We focused on research papers written in English --although many of these works study PLMs trained in languages different than English. We were also interested in published work that had been peer-reviewed, i.e. journal and conference research papers.

Our initial keyword search yielded a large amount of articles --around 5,580--, which was not a manageable number to handle and annotate given our existing resources. Furthermore, we verified that the majority of the recovered works were not related to our topic of interest. In order to narrow down our list to a more sensible number, and to choose a list of published work that was still relevant for our purposes, we followed a series of steps that are explained below.

First, we introduced a minimum impact threshold that consisted of excluding from our list any article that was cited less than three times by other works. We then held in a separate list all articles that were hosted in the preprint publication service ArXiv.org --1,040 of them--, since many of their provided research papers are potentially not peer-reviewed. This list underwent a semi-supervised review effort, which we will describe later in this section.

With the remaining articles from non-ArXiv sources, we then relied on the ranking score provided by Google Scholar, which measures the relevance of a given article against its set of search keywords. We used this information as a guideline to select, for each year of publication in our survey, the first 100 non-ArXiv articles in the list, an order that was specified by its ranking in Google Scholar. We ended up with 600 potential candidate articles for the timeframe between 2018 and 2023, and an additional 85 articles from January to October 2024\footnote{Proportionally, we retrieved 85\% of the articles, given it was a shorter time period}.

Returning to the held-out list of articles from ArXiv.org, we could not simply exclude these articles from our consideration, as many relevant sources might be found within this database. Despite ArXiv in itself not being peer-reviewed, Google Scholar provides in many cases, by default, an ArXiv-based version of research papers that are also published in credited conferences and journals. As a result, with our held-out list of ArXiv articles, we also performed a search within the ACL Anthology database\footnote{\url{https://aclanthology.org/}} to see which of those articles had been published there as well. The ACL anthology was adopted because it represents a relevant source of research work in the area of computational linguistics, comprising important venues such as ACL, CL, COLING, ConLL, EMNLP, IJCNLP, LREC, NAACL, RANLP or TACL, among others. Once duplicates were removed, we retrieved a set of 557 articles in the ArXiv list that were also part of the ACL Anthology database. Given that it was a large number -- especially if it were to be merged with the other 685 articles selected prior--, we followed a similar criterion as in our non-ArXiv collection of documents: we selected all articles in this list whose Google Scholar ranking score place them in the top hundred position in their publication year. This led us to choose a total of 191 articles from this source.

\subsection{Screening and inclusion}

We divided our initial collection of 876 articles into equally-sized sets to be validated by six human annotators. In this step, each annotator was tasked with manually revising the title and abstract of each work --and the complete text of the article only in case of need or doubt--, in order to quickly discard any possible articles that were not of our interest. The reviewers were posed with the following \textit{selection question} in order to choose among their assigned articles:

\begin{quote}
    ``Does the paper deal with the identification, analysis, and/or quantification of the \textit{linguistic knowledge} (e.g., morphology, syntax, semantics) that might be codified inside the internal structures of a Pre-trained Language Model (PLM)?''
\end{quote}

Each list was revised by two annotators, who were asked to answer the selection question with one on these options: ``Yes'', ``No'' or ``Doubt''. In case of ``Yes''/``No'' mismatches or articles marked as ``Doubt'' by both annotators, a third independent annotator was involved to arrive at a final decision. Other than that, cases of  ``Yes''/``Doubt'' mismatches were marked as ``Yes'', and ``No''/``Doubt'' cases were marked as ``No''.
Before the involvement of a third annotator, the average inter-annotator agreement rate was found to be 0.56 in Cohen's kappa coefficient, which means a ``moderate'' agreement rate according to~\citet{landiskoch}. Once the annotation clashes were promptly solved, we were able to narrow down our list to 277 articles.

\subsection{Paper analysis}

With this final list, we involved a human annotation effort again to perform a more in-depth analysis of each paper, in order to achieve a comprehensive understanding of their contents and ideas. This was accomplished by a thorough reading of each paper and its classification across the following dimensions:

\begin{itemize}
    \item Addressed linguistic level(s): phonological, morphological, syntactic, (lexico-)semantic, pragmatics and discourse. 
    \item Studied linguistic phenomena (e.g., anaphora resolution, concordance, etc.).
    \item Comparison with psycholinguistics theories\footnote{Psycholinguistics is a discipline that attempts to discover the psychological processes going on in a speaker’s brain that allows them to talk a language. The works that follow this perspective of study attempt to borrow some knowledge, tools, etc. from psycholinguistic studies in humans, in order to to see how feasible it is to apply them to the study of PLMs as if they were human subjects.}.
    \item NLP-like tasks mentioned, if applicable (PoS tagging, dependency parsing, etc.).
    \item Language(s) being analyzed in each model.
    \item Model(s) analyzed. Is it a multilingual model being analyzed?.
    \item Methods used for linguistic information discovery/analysis (e.g. probing, analysis of embeddings, ablation, etc.).
    \item Technical work available (e.g. framework, datasets...).
    \item Does it support the idea of PLMs effectively capturing linguistic knowledge? Or the complete opposite? Or something in between (‘it does \textit{X} well, but not \textit{Y} that good’…)?
    \item Does it use only the last layer embeddings or just the output of the model for performing its experiments?\footnote{A paper that was labeled as a positive in this category was to be excluded from our list. This is because we considered that such studies do their analysis in a `black-box' style, while not truly dissecting the internal parameters of the Transformer.}
  
\end{itemize}

Following this categorization, we also managed to discard another series of unrelated research work that had not been detected in prior steps. Our final list of work comprised 160 articles relevant to our search.

\pagebreak

\section{A brief overview of the Transformer architecture elements}
\label{transformer-elements-definition}

\begin{wrapfigure}{r}{0.54\textwidth}
  \centering
  \includegraphics[width=0.54\textwidth]{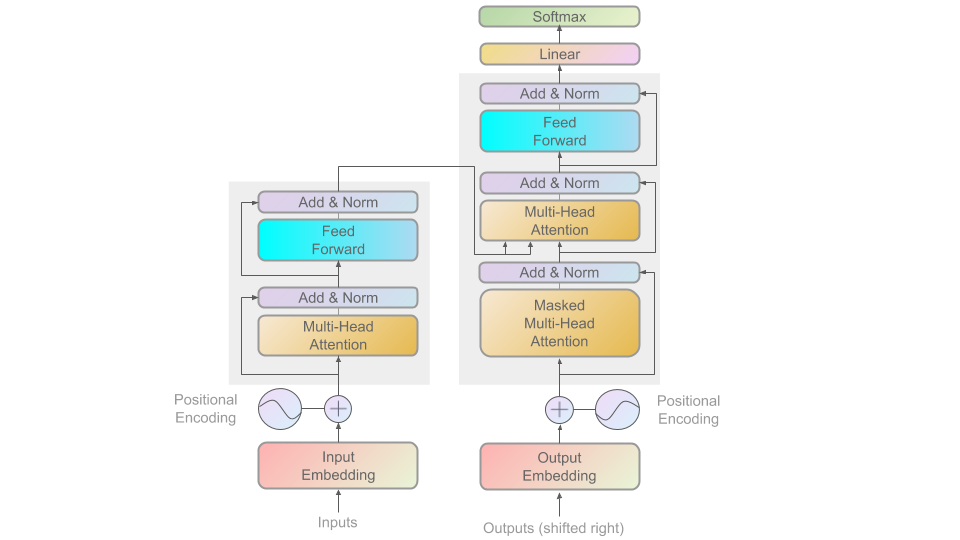}
  \caption{Simplified scheme of the Transformer architecture, consisting of two interconnected encoder (left) and decoder (right) modules, each comprised of a single layer (surrounding gray rectangle). 
  [Source of image: \url{https://github.com/dair-ai/ml-visuals}]}
  \Description{An example of the components in the Transformer architecture, explained via a simple model.}
  \label{fig:transformer-elements-figure}
\end{wrapfigure}

In this section, we provide a small overview of the different elements inside the Transformer architecture, to better follow the reminder of this paper.
A Transformer is a type of neural network, consisting of a series of individual \textit{neurons} that control the flow of information within the network, although with its own organization. The Transformer was originally designed to process text data --previously converted to vectors, as described later-- and return a series of other vectors, containing rich information encoded by the model, which are then transformed to a desired output --e.g. other text, representing an answer, in chatbot-based applications, or a label (True/False) in a text classification task. A Transformer is comprised of several interconnected modules called \textit{layers}, which are identical one to the other, and which repeatedly process input text data. The output of one layer serves as the input to the next one, until the end of the model is reached. The number of layers can change per model and is determined by their creators.\footnote{For instance, some models such as BERT or RoBERTa feature 12 layers, whereas other models can feature more layers.}

A preliminary step, called \textit{tokenization}, is needed before processing the information by the different layers. In this step, the input text is split into \textit{tokens}, which do not necessarily correspond to individual words. In the case of BERT~\citep{BERT}, tokenization is performed at the level of subwords, which separates texts by individual words but also splits some commonly recurring forms --e.g. the ending "-ing" in many English-based verbs. As of today, however, the most commonly used tokenization scheme is called Byte-Pair Encoding (BPE), which is used by RoBERTa~\citep{roberta} and all GPT-based models --including ChatGPT~\citep{OpenAI}. BPE divides texts not based on linguistic cues --such as individual words or morphemes--, but based on text compression artifacts.
Then, the tokenized text reaches the first layer or the Transformer (the  \textit{embedding layer}), which has a different implementation and purpose than the rest of layers. This initial layer converts all the input tokens to a series of vectors, each with an initial fixed value, via a look-up table, in order to be handled by the rest of the architecture.

The rest of individual layers take as input a series of vectors, one per token, and contextualizes each in regards with the text they are in. The contextualized vectors output by each layer are called \textit{contextual embeddings}.\footnote{This is a name borrowed from the tradition of word embeddings, retroactively called `static embeddings', which is a popular distributional method for representing words --e.g.~\citep{word2vec}-- in NLP research.} Each layer accepts a maximum of $n$ contextual embeddings, and outputs at its other end the same number of contextual embeddings. The number of accepted contextual embeddings per layer differs in each model.
A layer itself is not a monolithic component, and is internally composed of several elements (see Figure~\ref{fig:transformer-elements-figure}):

\begin{enumerate}
    \item The \textit{self-multihead attention} layer, a component that takes each input embedding and compares its value against that of the other contextualized embeddings from the rest of the text. These attention heads contain a series of learnable weight matrices, against which the embedding vectors are multiplied and from which they obtain their contextualized representation. The values of these matrices are determined during training of the model.
    \item A regular \textit{feed-forward layer}, which updates the output embeddings of the self-multihead attention layer through two linear transformations: the first one expands the dimension and the second one projects it back to the original dimension.
\end{enumerate}

When a specific work broadly mentions \textit{lower}, \textit{middle} or \textit{upper} layers regarding the processing of a specific phenomenon, this can correspond to a set of layers depending on the model. For the case of BERT-like models with 12 layers, lower layers would roughly correspond to layers 1-3, middle layers to layers 4-7 and upper layers to layers 8-12.

Furthermore, a Transformer-based model can be (i) an \textit{encoder} model, also referred as \textit{autoencoding}, such as BERT~\citep{BERT} or RoBERTa~\citep{roberta}; (ii) a \textit{decoder} model, also known as \textit{autoregressive}, like GPT-based models; or (iii) an \textit{encoder-decoder} such as the original Transformer or BART~\citep{lewis-etal-2020-bart}. The difference between them in how they handle the self-attention in the above mentioned layer: whereas an encoder considers all tokens in the sentence to calculate the contextual embeddings, a decoder generates the contextual embedding of a token considering only the tokens situated to its left. This makes the latter more suitable for text generation tasks, hence its wide adoption in chatbot-based applications. The encoder-decoder models combines both, being suitable for sequence-to-sequence tasks such as machine translation.

\section{Quantitative analysis}
\label{quantitative-analysis}

In this section we present a series of observations found across our selected research work. These are mostly quantitative and aim to provide a perspective on the current development and tendencies in linguistic interpretability studies.

\subsection{Languages and studied model architectures}

We have observed a tendency in the studies of linguistic interpretability in PLMs to focus mostly on models trained in English. Of our survey of 160 studies, 145 of them study language models in English, whether in a monolingual setting (103 studies) or alongside other languages (42 studies). Despite this tendency for English-centric studies --a common pitfall in the development of NLP-based technologies--, we have observed a big number of studies in other languages, which means the study of linguistic interpretability is steadily opening up for other language families. Minoritary languages, however, are understudied, something that could be a reflection of their overall lack of representation in the PLM world. In Table~\ref{tab:langs_count} we see a collection of the languages studied in our survey.

Regarding the analyzed PLM architecture type, which we can see in Table~\ref{tab:models_arch_analyzed}, our list of studies mostly rely on the study of BERT-based monolingual models, alongside RoBERTa monolingual models. Multilingual BERT is also highly studied. We see no studies in this list, however, on more modern models such as Llama or BLOOM, and there are very few studies on decoder-based architectures --despite their growing popularity.\footnote{Table~\ref{tab:models_arch_analyzed} only includes model architectures that have been studied by at least two research papers in our survey.}

This fact, however, is to be expected, owing to our focus in this survey on interpretability methodologies that rely on internal parameters.
Interpretability techniques that make use of internal parameters have become increasingly unaffordable in current state-of-the-art PLMs~\cite{Explainability_for_Large_Language_Models_A_Survey}, due to their ever-growing sizes as well as the closed nature of some of the models --such as the GPT family after the release of GPT-3. This has led to the adoption of prompting as a generalized technique for language tasks, owing to its low technical overhead. This, however, also comes at the expense of lack of transparency and loss of potential interpretability on the internal calculations involved in the final results. In this regard, BERT and RoBERTa, even if aging, are models that still show competitive performance in language tasks --though not on par with more contemporary models--, and their more modest sizes means that they can still be used with interpretability techniques that explore models from within. 
This leaves room in the future for the development of studies that perform this type of linguistic knowledge analysis --i.e. layerwise-- in more recent, decoder-based models. 

\begin{table*}
    \centering
    \caption{Number and list of languages in PLMs analyzed in our survey}
    \label{tab:langs_count}
    \small
    \begin{tabular}{ccllllllll} \hline 

    \textbf{Language} & \textbf{\#} & \textbf{Language} & \textbf{\#} & \textbf{Language} & \textbf{\#} & \textbf{Language} & \textbf{\#} & \textbf{Language} & \textbf{\#} \\ \hline
    
         English& 145& Basque
&10&Dutch&5&Norwegian&3&Serbian&2\\  
         German
& 21& Hebrew
& 8& Estonian& 5& Slovak&3& Vietnamese&2\\  
 French& 20& Polish& 8& Greek& 5&Ukrainian&3& Armenian&1\\  
 Russian& 20& Swedish& 8& Latvian& 5&Urdu&3& Filipino&1\\  
 Finnish& 17& Arabic& 7& Bulgarian& 4& Afrikaans&2& Kannada&1\\  
 Turkish&17& Japanese& 7& Croatian& 4&Albanian&2& Macedonian&1\\  
 Chinese&14& Portuguese& 7&  Danish& 4& Belarusian&2& Mandarin Chinese&1\\  
 Italian& 14& Farsi/Persian& 6&Marathi&4& Catalan&2&Nepali&1\\  
 Spanish& 14& Hindi& 6& Romanian& 4&Icelandic&2&Tagalog&1\\  
 Czech& 11& Indonesian&6& Slovenian&4&Irish&2&Telugu&1\\  
  Korean& 11&  Tamil& 6& Lithuanian&3&Latin&2& Yoruba&1\\ \hline
    \end{tabular}
\end{table*}

\begin{table*}
    \centering
    \caption{Main model architectures studied in this survey}
    \label{tab:models_arch_analyzed}
    \small
    \begin{tabular}{cccc} \hline 
    \textbf{Model} & \textbf{\#} & \textbf{Model} & \textbf{\#} \\ \hline
    
         BERT + monolingual variants
&  122&  DeBERTa&  4
 \\  
         RoBERTa + monolingual variants
&  39 
&  Glove (static embeddings)&  4

 \\  
         mBERT&  38
&  GPT&  4

 \\  
         GPT-2 + monolingual variants &  13&  XLM
&  3
 \\  
         XLNet&  13
 & GPT-2-XL &  2\\  
         ELMO (non-Transformer)&  12
& InferSent (non-Transformer)  & 2 
\\  
         XLM-R&  11 
& mBART  & 2 \\  
         ALBERT&  6
&  miniLM &  2
\\  
DistilBERT&  6
&  T5
&  2
\\  
ELECTRA&  6&  TransformersXL &  2\\ \hline
    \end{tabular}
\end{table*}

\subsection{Methods for linguistic discovery}

The interpretability of PLMs is subject to a plethora of different techniques used in the attempt to explain the inner workings of these models. As such, each research paper in this survey can present their own language discovery methodology and overall conclusions. Even if an article reuses an existing methodology from another research, they might apply some modifications in their attempt to uncover some hidden internal functionality of the studied models.

Despite this great diversity of methods, some surveys in the literature have attempted to perform a broad categorization of the type of interpretability techniques commonly used in the analysis of Transformer-based models. Among them we can mention the ones presented by~\cite{Explainability_for_Large_Language_Models_A_Survey, XAI3, XAI4, XAI2}.
These existing taxonomies for interpretability of language models, while useful, proved difficult to apply as-is to our list of studies. 
For the purposes of our survey, we have devised an alternative classification of discovery methods that, while borrowing cues from the already mentioned works, has also been simplified and adapted to our specific use case. This list of methods is not meant to be exhaustive neither represent a fixed taxonomy, but rather provide a broad categorization of the overall methods that are in use across the papers in our survey:\footnote{Our proposed taxonomy is also compatible with those provided in other research works, e.g.~\citep{Explainability_for_Large_Language_Models_A_Survey}, and complement them for the specific case of linguistic knowledge discovery in PLMs.}

\begin{enumerate}
     \item \textit{Feature attribution methods}: Techniques that quantitatively measure the relevance of each input feature (e.g., tokens, phrases, text spans) with respect to a model’s final prediction or outcome. Methodologically, this measurement is \textbf{always} performed via the calculation of a \textit{relevance score}. As an example, a non-Transformer specific framework for feature attribution would be the model-agnostic tool LIME~\citep{LIME}.
     Possible techniques might include \textit{perturbation}, which alter input features to observe changes in model output (e.g., change the order of the words, omit some words, etc.) --e.g.~\citep{9517593e2fca883a1391c8d8068f4426fa7663cab61074fe7e784297edd020b7}; \textit{gradient-based methods}, which compute per-token importance scores using the partial derivatives of the output with respect to each input dimension --e.g.~\citep{additional_108}; \textit{surrogate models} --e.g.~\citep{82436c5c586808f7e49bec663f288b6637ecd907f5086aac3132cb16c3546545}--, which approximate the “black-box” PLM model with more interpretable models –i.e. “white-box”– such as decision trees, linear models, etc.
    
    \item \textit{Example-based explanation}: Methods that use specific input examples, which are carefully chosen --e.g. sentences containing specific linguistic phenomena--, in order to study the behavior of a model. Unlike feature attribution methods, example-based explanations do not quantitatively measure the impact of each individual input feature through a relevance score; instead, these methods leverage full-fledged text-based examples and assess how a model’s output changes with those different examples.\footnote{A dataset containing sentences with changed word order can be instances of example-based explanations when these sentences are ultimately not subjected to the calculation of a relevance score, but passed-in as-is to the model. However, if a relevance score is involved in this process, then those perturbed sentences will be part of a feature attribution method. The difference, in this sense, is mostly methodological.} 
    Possible techniques might include \textit{adversarial examples} (introduction of small, imperceptible alterations to input examples, with no discernible linguistic or semantic disruption apparent to human observers, but to which the underlying Transformer model might struggle at) --e.g.~\citep{20a81fe3cc98aac1cc4b4be98d5f23ce2a97c4d366641647641d3ad26abdbbca}-- or \textit{counterfactuals} (akin to adversarial examples, but with much major disruptions performed internally at the level of the contextual embeddings) --e.g.~\citep{0a8f4755cc0cdbfa5ba0f95dfda7061bd32f36a8c81e586fffed249ec106da83}.

    \item \textit{Analysis of architectural elements}: Some fixed internal components of the Transformer architecture (such as attention heads, neuron circuits, feed forward layers, etc) or intermediate representations (contextual embeddings, logits, etc.) are studied directly to understand their role in the obtention of the model’s predictions or outcomes. Possible techniques include \textit{visualization of attention patterns} (by means of bipartite graphs or heatmaps for specific inputs, in order to visualize how the attention mechanism tends to attend to the most relevant tokens) --e.g.~\citep{additional_09}; \textit{tracking of attention weights} (also explores the attention mechanism but does so quantitatively) --e.g.~\citep{additional_178}; \textit{analysis of the feedforward network layers} and what sort of linguistic  knowledge they encode; \textit{neuron activation explanation} (examines individual neurons that seem to be important for specific linguistic phenomena); \textit{embeddings analysis}\footnote{This category partially overlaps with example-based explanations, as they constitute specific examples being analyzed, but they are included here because they refer to the analysis of internal Transformer architectural components.} (where contextual embeddings are analyzed directly, either with statistical methods or visualizing their internal geometry) --e.g.~\citep{ff69f49e3a5e82a147575bc0dfcc94bcf3848fc9c90595829952ec3bbb706041}; \textit{model pruning}, i.e., selectively removing some components of a PLM to check the impact of their erasure on the final results --e.g.~\citep{20a81fe3cc98aac1cc4b4be98d5f23ce2a97c4d366641647641d3ad26abdbbca}.
     
    \item \textit{Probing}: Broad specification used to refer to methods and/or architectures that leverage a model’s intermediate parameters –-e.g. contextual vectors–- to learn a global representation that resembles linguistic knowledge, and which can be used as direct or indirect proof of a PLM’s knowledge of language~\cite{belinkov-glass-2019-analysis}. This representation is usually learnt by a dedicated external classifier, which is trained on a set of linguistically-annotated training datasets\footnote{Although many probing experiments make use of specific datasets to train a probe –-which could lead to confusion with example-based explanation–-, the purpose of those datasets is to help learn a global representation for the probe, and \textbf{not} to analyze those specific examples -–this is the reason why they belong to a different category. On the other hand, in the scope of our taxonomy, if a probing dataset being used contains specific phenomena –-e.g. changed word order–-, then we consider it to overlap with example-based explanations.}, which can consist of any sort of internal parameters, whether contextual embeddings from intermediate layers or attention weights. The classifier is meant to solve a task called a ‘probing task’ during its training. If the final representation learnt by a probe shows consistent results for new, unseen representations, then the source PLM is said to possess the analyzed linguistic capability.
    Probes are architecturally very different one from the other, with a common recommendation across many works to make them as lightweight as possible --although there are other studies~\cite{additional_54, additional_80} that promote the opposite idea.
    A potential risk of probing is that of a classifier probe simply memorizing how to perform an end task instead of diagnosing it in a target PLM.\footnote{For example, a syntactic probe might simply be learning how to perform parsing from scratch on the contextual embeddings it has been provided. This is one of the reasons why many works in interpretability vouch for modestly-sized probes, so as to give these models as little room as possible to learn their task and avoid --or minimize-- this issue.} Several solutions have been proposed to demonstrate that memorization does not take place, such as the training of a set of auxiliary probes --e.g. `control tasks'~\citep{hewitt-liang-control-probes}-- deployed alongside some main probes. 

\end{enumerate}

We should warn the reader on our use of the word `probing' in our survey, which may differ from other research papers found outside this work. These other studies may tend to make an alternative use of this term, in which it is referred to as a synonym of the general task of discovering knowledge in PLMs --even if these works do not use probing techniques per se. In the context of this survey, we refer to the term `probing' only in papers where probes --i.e. classifier modules that discover linguistic knowledge within intermediate PLM representations-- are actually in use for unearthing linguistic knowledge in PLMs. Our definition of probe is also strict, in that it must consist of an external classifier module that learns a linguistic feature from intermediate PLM representations.

Despite the different methods for linguistic discovery in use, we have found that many papers in our list --122 in total-- make use of the probing methodology, either exclusively or in combination with other methods. In many cases, we have found that papers in our survey combine probing with other types of methodologies.\footnote{For instance, \citet{additional_122} make use of classifier probes but the input to these classifiers is perturbed text, which generally falls under example-based explanation techniques.} The use of combined methodologies lead to more diverse and rich results. In Table~\ref{tab:methods_count} we can see the number of research works that utilize each of the different interpretability techniques listed above.

\begin{table}
    \centering
    \caption{Number of times each method for linguistic discovery in PLMs is used across our survey}
    \label{tab:methods_count}
    \small
    \begin{tabular}{ccccc} \hline 
         &  \textbf{Feature attribution} &  \textbf{Example-based} &  \textbf{Analysis of elements} & \textbf{Probing} \\ \hline 
         \# & 16 & 16 & 58 & 117 \\ \hline
    \end{tabular}
\end{table}

\subsection{Linguistic disciplines and specific linguistic phenomena}

We have analyzed the global linguistic level addressed by the studies in our survey, identified by the following classical  disciplines in Linguistics: phonology, morphology, syntax, semantics (and lexicosemantics) and discourse. Many studies do not focus in one but several disciplines at the same time. We found that articles on syntax are the great majority (113 articles), closely followed by semantics/lexicosemantics (82 articles). This means many researchers are interested in studying either the hierarchical syntactic structure encoded by Transformer models, or their knowledge of semantics. Morphology has a lesser number of retrieved articles (38 in total), while we found 6 articles for discourse and we only managed to retrieve a single article on pragmatics. No article was found on the area of phonology, but this is to be expected given it is a discipline that is better handled by multimodal LMs --which we do not analyze here. We also include these numbers in Table~\ref{tab:ling_levels_count}, as well as the references to each of the studied works. Owing to space limitations, we have prepared a more comprehensive list of the results of this survey
in supplemental material provided.

While some papers delve into the study of a linguistic discipline --e.g. syntax-- as a whole, a number of them also analyze specific linguistic phenomena and how they are encoded in PLMs. Such studied phenomena include null-subject~\citep{6c0bcfd8ee5a86207dcb0ac639539437b705ee2b36ce47cc5e645d9d6a521200, 788f3f1df1a8c3daa1e4bbfdf6afbe350660000ad607b64a238cd4946bbc63f5}, agreement (subject-verb: ~\citep{additional_07, 788f3f1df1a8c3daa1e4bbfdf6afbe350660000ad607b64a238cd4946bbc63f5, ca72b08d8fc0d2b43cc951a6ea692310a3e6dc02f927d39eeffaead647c73b28, additional_129, additional_147, additional_154}; noun-verb:~\citep{additional_124}; agreement violations:~\citep{3d84df6267a2e73cd70576d10a8b6855250044ab8eea8f382060faeec4a09713}), negation~\citep{6d5cc37605e9d476ec1fe06906f83e72c4ca0fc45caaea21ecae1aebbcd6c50e, 9741c619a3f09f7a316e0907909293b33562493c6ed4f5c033078d710a614ec7, 8a88606731bd34583429af7deac06aa5de993adb02d231fad354ac7ad58e91dc}, grammatical number/gender/tense~\citep{additional_162, additional_161, aba65a2f5df012a7f434e275f03d8cbd63c6fb56fc12a68d1dfcf877ae4bc7a3, additional_117, 24bfbc52ddfe90eed7562e45f4764630f0befcd1825d6e995f459dbe3b59f23e, bb77fe70885ba02cbc438e9ba43185fcf11a46359ee71738a2e39975ee07fc30, 9517593e2fca883a1391c8d8068f4426fa7663cab61074fe7e784297edd020b7}, compounds~\citep{additional_189, 402715498773cde99577331d0d88d500cf48e07f23d0f204e335371946f6b99b}, correference~\citep{611256713eaca002037294e94ef06a1790d87973ee0556976b768c74bdba0118, additional_04, 0ea4eaa575b9fe22c52fe5eccb54c7449f1f24c2652c3d89ae8e4b96eb5d6720, bb1665b9f77e8a7fb0cd25dcd438d563fd436dd5f54957c42fd9e3e676593855}, metaphors~\citep{additional_176}, garden-path sentences~\citep{172f1fc1d8e4bbaa92585f9e87a7b3b0df46038506183559c3a6ef42cf929002}, idioms~\citep{730b1312face40c86e1bb39615e5032b054829cbf9d47753d186adac061a46ff} (their non-compositionality in~\citep{additional_178}), etc. Some studies even encompass the study of a great number of linguistic phenomena at the same time~\citep{additional_64, 8a88606731bd34583429af7deac06aa5de993adb02d231fad354ac7ad58e91dc, 24bfbc52ddfe90eed7562e45f4764630f0befcd1825d6e995f459dbe3b59f23e, additional_117, aba65a2f5df012a7f434e275f03d8cbd63c6fb56fc12a68d1dfcf877ae4bc7a3, additional_05}, mostly if these phenomena have been annotated as part of a preexisting large-scale linguistic dataset.

Interestingly, there is a line of work that does not study traditional linguistic concepts, but aims instead to measure the proficiency of PLMs in several classical NLP tasks that are highly related to language, such as Natural Language Inference (NLI), Named Entity Recognition (NER) or Part-of-Speech (PoS) tagging --e.g.~\citep{additional_138, additional_54, additional_64, additional_59, a8c3c25f32aa64fd200681812f190528981a14639d8a2099497d7a1217c3cede, 198f16c09501e24c2d3c71612c23c022fbb5d428f882da4e59e303fc14c4ac21}. In these works the aforementioned tasks are sometimes equated to linguistic knowledge. This represents a methodological mismatch between the notions of traditional linguistics and that of computational linguistics in the study of this type of knowledge in PLMs --however, in some papers, these NLP tasks are mentioned as support tools for the discovery of traditional linguistic concepts in models.\footnote{For example, \citet{additional_93} rely on NLI techniques for the discovery of syntactic knowledge in PLMs.} The same situation happens with Universal Dependencies (UD) notations, which are also present in some papers across our list --e.g.~\citep{additional_64, additional_59, ae4d7d191fc19095d3f4fd18b30db31790711ace2969f273e63207f1b83cf823}.

\begin{table}
    \centering
    \caption{Linguistic disciplines found in our survey and the studies in our list of analyzed research works}
    \label{tab:ling_levels_count}
    \small
    \begin{tabular}{>{\centering\arraybackslash}p{2cm}>{\centering\arraybackslash}p{10cm}c}
    \hline
 \textbf{Discipline} & \textbf{Studied works} & \textbf{No. of works} \\
    \hline
         Syntax& \citep{additional_110, e0c6839372ecd8ff0f39cc8f5e1615da95551baab4bf7e95611826308fbd74d7, 788f3f1df1a8c3daa1e4bbfdf6afbe350660000ad607b64a238cd4946bbc63f5, 8bdef06d2dfc5d96f9d10858c17232603ebb24103b3e50add08435b745bd471d, 9517593e2fca883a1391c8d8068f4426fa7663cab61074fe7e784297edd020b7, additional_159, de14b3483b08a04f231f0908122bf595363d9bf8fb31c2840302d2d56a3be625, additional_07, b232ac3efe94d7a4049c8325bab09693ea24cfe2c311018c755009bfb6355d24, d4cdf3cee4798cb5ccbf020fab77ea3eead1d432fe8e3873ec5731f2f8bf7053, additional_144, f427a07607385b6e699bf4ef714b60cf555577e542477eff9f35d2a08d626518, 084126497fcc2eab7622af1eec6ab948fd5dc362349adba3d932e617a978664e, 52e3a3ee7d5e53eab5b9be8ae2c5370829669971a9a4096180652f08dbec520d, additional_90, 172f1fc1d8e4bbaa92585f9e87a7b3b0df46038506183559c3a6ef42cf929002, additional_177, 6bde56f74fa302633de19099fc56d2a3247291067a76b5473c027460d6d2eac5, 82184cce63c8cc9d33442c9097ae89d90ccd17290b20b180777a203ca19fc5ef, 5dae30a5a97b2a1cb7e3b3f1b4ebfa86895991b1adaa6a1c9d8b4f79c7f73ae8, f284d021afd9d17abe71095a7c2879adaf732a37b906048c9796d94deb4eed23, additional_130, 099659e8eec03507a3c4e46c25fb3566eaf52ce9e4f995024d0ec341730da708, f79f09229def9cf277ff743fdb8c86896e0d5f9b3206423953cd48adeb6c809b, 2ab1cd858abb87043053d4aa1c3d8fcc9ab7381f1788ee5e77213ad54e32037d, 4ae9f84833b4deb3ea3aba5a62bfab93b0d6c27c440be7f2be4ec98772c41f0a, additional_161, additional_169, additional_55, 1bce61f930980f7a779b24aeb82bd4deb4cadfeb316b5253fd3793dda36e29eb, additional_74, f980b6bb7db7c57e5d4c640e3cf54117d8a2e65667318ee8a2ac5e9320b38b86, additional_162, additional_124, ca72b08d8fc0d2b43cc951a6ea692310a3e6dc02f927d39eeffaead647c73b28, additional_81, additional_108, additional_156, ffdba41698f72c61da951d9d03e1f94273e733b785d08a1a40d448c68facb163, 9ee3dcd70c9bf67f6929ab7ae1d6844c94c9c398bb313c855adaa0e4bd87cfae, additional_150, additional_26, additional_129, additional_168, a8c3c25f32aa64fd200681812f190528981a14639d8a2099497d7a1217c3cede, 7744824c4cd439e2c16af583034e3da118e2cd3fb180709cc1c00ed99ca07be6, additional_08, 0ea4eaa575b9fe22c52fe5eccb54c7449f1f24c2652c3d89ae8e4b96eb5d6720, additional_171, additional_147, additional_112, additional_165, 41128f3f273ba991eaff22672cb2714ea9e471d273d3c6f4dcfa590d5e173fdc, af421422e4bec0fa1e8a222f10006fb390c7483e9843b5dcf7e84912ec062477, additional_122, additional_173, 03f3b85dff8f1061859ae1f41c4e6c31b25dda098fe98d3f8443a008c9e1de3f, additional_160, 6d5cc37605e9d476ec1fe06906f83e72c4ca0fc45caaea21ecae1aebbcd6c50e, additional_163, additional_87, ae4d7d191fc19095d3f4fd18b30db31790711ace2969f273e63207f1b83cf823, ab5e694285556e053068e170ef7865a4a3997f4133b60adb6410048776d74eef, 70aa455084d7f33f92386a31831a20f9c348693ed132794c7194b47b78e4de1a, additional_80, additional_142, additional_154, a420264b61d7e28e6a709b51b9146c1b13c19d45655e8a2f9363194f22a4be4e, additional_93, additional_05, additional_115, additional_59, 6c0bcfd8ee5a86207dcb0ac639539437b705ee2b36ce47cc5e645d9d6a521200, 5cabfa60fe1197579bf66eb101e3d6110f007839095bc942706e6cdaf51eb843, e6c453545e421dd0d4a75b0a7b418660449471b7277016e71d042b71416aeae2, 4f5d09b2287c7510fed7b99addae1ad41b7ba16bf061509b0bd92b310e38b1d1, b7ec8a05b964b2a7bf7b0016b96e3c83867e03b84978fbe4356572bde8a3f561, additional_114, 2e34ff32b63e31f6c144010143a0a51be955cb64583653296dfa868d529e6b1d, a367657d6ae086dac6985fdb6ccb18dd6c0e8727e4b3c5903b5f362cc46dc7ca, aba65a2f5df012a7f434e275f03d8cbd63c6fb56fc12a68d1dfcf877ae4bc7a3, 867bde36723c79dc106d857c379bcdf584fb5f04a18b34d034ff3da4133d59f7, additional_117, 9741c619a3f09f7a316e0907909293b33562493c6ed4f5c033078d710a614ec7, additional_107, additional_09, 292e02b264e2b608b7f94ceb4e3ba02c8c39420609d170ccc537f1e2f7bbc4b5, 0a8f4755cc0cdbfa5ba0f95dfda7061bd32f36a8c81e586fffed249ec106da83, additional_58, eef9d3252ff314123f8ef485036e676bd933b3a0d0647e23992e5522eec1b178, 8a88606731bd34583429af7deac06aa5de993adb02d231fad354ac7ad58e91dc, 7bca6b86363d9c8ab41964b3f2c90f9e0684e1e0c96e14e3617cefc6ebbf315f, 24bfbc52ddfe90eed7562e45f4764630f0befcd1825d6e995f459dbe3b59f23e, additional_186, additional_138, additional_64, additional_157, additional_54, additional_116, 6a9679421f4cc72dbb987a7acbc37cf4863a5dec5d1547db016b8f6215bc8cfc, additional_92, additional_68, dee4ee9db0e8ad44bfb13b1493091a6206e87cda9327ec9a7900a5ae1bd2e9d7, additional_145, additional_29, additional_21, 611256713eaca002037294e94ef06a1790d87973ee0556976b768c74bdba0118, 39e608673194e619191dbda4a5f348a8af0dca9b6603e901c28598351c10afb5, 3d84df6267a2e73cd70576d10a8b6855250044ab8eea8f382060faeec4a09713, additional_125, additional_77, 198f16c09501e24c2d3c71612c23c022fbb5d428f882da4e59e303fc14c4ac21, 60388e31f4f00a3fa25c01e3e84d9d5253770a3287a25fee2014b744f5ff41c3, c150a691ac9f33ab1c91126b1f53eabc2ce350893235c59b1c34187fdf2d6382, 9bf8105e551b32bbfe1c816f14e7d71c06ea805e1635c577ed125a315c542f84, additional_04, twentyfour_04} &113\\ \hline
         (Lexico) Semantics& \citep{e0c6839372ecd8ff0f39cc8f5e1615da95551baab4bf7e95611826308fbd74d7, 730b1312face40c86e1bb39615e5032b054829cbf9d47753d186adac061a46ff, 5b0ad7a02f7f5dde79c6b9fd147841d6e22ad9cf10873ceac3777e4e565897b7, 82436c5c586808f7e49bec663f288b6637ecd907f5086aac3132cb16c3546545, d0eb7251a950c57432892c84c25caa8e8a936348d985236d352da233dfdd18f8, ac6ae2754b9211ea8e715a6c6a8d82ce222c95d1c69fa1f283c5a675809fbd03, 8bdef06d2dfc5d96f9d10858c17232603ebb24103b3e50add08435b745bd471d, additional_178, de14b3483b08a04f231f0908122bf595363d9bf8fb31c2840302d2d56a3be625, 2d609b540a949ba868404a5a68672251f99dc620c0d3b08d32a6a1140553e53b, d4cdf3cee4798cb5ccbf020fab77ea3eead1d432fe8e3873ec5731f2f8bf7053, additional_144, additional_90, 77b165db81953eb8217e121eb7ae050919a23cab7146ee76c4a975fb9f0072b7, additional_177, 6bde56f74fa302633de19099fc56d2a3247291067a76b5473c027460d6d2eac5, 5dae30a5a97b2a1cb7e3b3f1b4ebfa86895991b1adaa6a1c9d8b4f79c7f73ae8, additional_89, a58e1f06742805ce6fa99fc3bd0db86c2fc39b5c7404bea798342dc91285226f, 099659e8eec03507a3c4e46c25fb3566eaf52ce9e4f995024d0ec341730da708, 85a3e9443e0fdf06889804fe1bb78f12d800be6df32d2c3a40db0fabd08c3074, additional_124, c150031d36ba8c897fa8cc81b743700a854c824c42ae375bdb6baaa7365c6a74, additional_108, f6a9d2a449ba32f9e7e2fc1656d815ec2d7437b80bc2a2ef2f54ce4b414690bc, additional_156, 9ee3dcd70c9bf67f6929ab7ae1d6844c94c9c398bb313c855adaa0e4bd87cfae, additional_168, 7d129a1596d1adb62bc05a641d18f71f70d840c7faee2ab9a72f8c8ea7a9c8f1, 7744824c4cd439e2c16af583034e3da118e2cd3fb180709cc1c00ed99ca07be6, additional_08, 2263597f16afdc0b04e5c64cecd337e8f2728c07a6a3ee46b06502245e7e5857, c4b9ab20f2a2bf267e75987b9d9fa6ac2fe15ed0b9b90c0b65922c85136532b4, 0ea4eaa575b9fe22c52fe5eccb54c7449f1f24c2652c3d89ae8e4b96eb5d6720, additional_147, additional_112, ff69f49e3a5e82a147575bc0dfcc94bcf3848fc9c90595829952ec3bbb706041, additional_165, 41128f3f273ba991eaff22672cb2714ea9e471d273d3c6f4dcfa590d5e173fdc, af421422e4bec0fa1e8a222f10006fb390c7483e9843b5dcf7e84912ec062477, additional_160, 6d5cc37605e9d476ec1fe06906f83e72c4ca0fc45caaea21ecae1aebbcd6c50e, additional_163, additional_87, 2f0b8010aeb5143941f96f1f398fc123f41f59674b6bc65229d61b61be8d3171, additional_16, 402715498773cde99577331d0d88d500cf48e07f23d0f204e335371946f6b99b, additional_142, a420264b61d7e28e6a709b51b9146c1b13c19d45655e8a2f9363194f22a4be4e, b2250768b4c31c7eda0c454127dc520e16c8c2059646036966e0dcbf0ec21f54, additional_05, 5cabfa60fe1197579bf66eb101e3d6110f007839095bc942706e6cdaf51eb843, bb1665b9f77e8a7fb0cd25dcd438d563fd436dd5f54957c42fd9e3e676593855, e6c453545e421dd0d4a75b0a7b418660449471b7277016e71d042b71416aeae2, 930460b95d9a6e0cd3e1e8e7c55d3a5491b2ba2c927d01eb9cd16aacc47ef47c, b7ec8a05b964b2a7bf7b0016b96e3c83867e03b84978fbe4356572bde8a3f561, additional_114, 4a343f0a56c9dea29ff8484c457c3a894f5b8d37c85f407b2f7723c96ccc3494, aba65a2f5df012a7f434e275f03d8cbd63c6fb56fc12a68d1dfcf877ae4bc7a3, 867bde36723c79dc106d857c379bcdf584fb5f04a18b34d034ff3da4133d59f7, additional_176, 9741c619a3f09f7a316e0907909293b33562493c6ed4f5c033078d710a614ec7, 292e02b264e2b608b7f94ceb4e3ba02c8c39420609d170ccc537f1e2f7bbc4b5, eef9d3252ff314123f8ef485036e676bd933b3a0d0647e23992e5522eec1b178, 8a88606731bd34583429af7deac06aa5de993adb02d231fad354ac7ad58e91dc, additional_66, additional_186, 0534fb05dc4279aae56d158d629c14ba54c24b60b36068890e43f5bc0adfa210, d6b058a6ded9412680d7bd40d1831b7a6a3bb262c46ba283188871c6a3e4cda9, additional_64, additional_189, additional_92, additional_145, b8243b9e6ab5a5f1d7f0777ea2928465082d00215d5ca57bc58a16d68efca413, 4809831c190d9712231d7022e8d438c0e1b0cb666c169632fdc8c5f787e8d828, additional_29, additional_21, additional_77, 755a82b7a6fbf89197a027287f273eb7f4f88301aa62fc4510f6350c6989590b, 198f16c09501e24c2d3c71612c23c022fbb5d428f882da4e59e303fc14c4ac21, 60388e31f4f00a3fa25c01e3e84d9d5253770a3287a25fee2014b744f5ff41c3, c150a691ac9f33ab1c91126b1f53eabc2ce350893235c59b1c34187fdf2d6382, additional_61, additional_56, 9bf8105e551b32bbfe1c816f14e7d71c06ea805e1635c577ed125a315c542f84, twentyfour_04}  &82\\ \hline
         Morphology& \citep{additional_110, 9517593e2fca883a1391c8d8068f4426fa7663cab61074fe7e784297edd020b7, f427a07607385b6e699bf4ef714b60cf555577e542477eff9f35d2a08d626518, additional_75, 52e3a3ee7d5e53eab5b9be8ae2c5370829669971a9a4096180652f08dbec520d, additional_90, bb77fe70885ba02cbc438e9ba43185fcf11a46359ee71738a2e39975ee07fc30, additional_177, 6bde56f74fa302633de19099fc56d2a3247291067a76b5473c027460d6d2eac5, 5dae30a5a97b2a1cb7e3b3f1b4ebfa86895991b1adaa6a1c9d8b4f79c7f73ae8, f284d021afd9d17abe71095a7c2879adaf732a37b906048c9796d94deb4eed23, additional_161, 1bce61f930980f7a779b24aeb82bd4deb4cadfeb316b5253fd3793dda36e29eb, additional_162, additional_124, 9ee3dcd70c9bf67f6929ab7ae1d6844c94c9c398bb313c855adaa0e4bd87cfae, additional_168, additional_08, additional_165, 41128f3f273ba991eaff22672cb2714ea9e471d273d3c6f4dcfa590d5e173fdc, additional_173, 03f3b85dff8f1061859ae1f41c4e6c31b25dda098fe98d3f8443a008c9e1de3f, fd02effe5f3f2424aee1e7b6f4be2bdf4619419336b3b32077a706955b79ca81, 2f0b8010aeb5143941f96f1f398fc123f41f59674b6bc65229d61b61be8d3171, ae4d7d191fc19095d3f4fd18b30db31790711ace2969f273e63207f1b83cf823, ab5e694285556e053068e170ef7865a4a3997f4133b60adb6410048776d74eef, additional_80, 5cabfa60fe1197579bf66eb101e3d6110f007839095bc942706e6cdaf51eb843, 4f5d09b2287c7510fed7b99addae1ad41b7ba16bf061509b0bd92b310e38b1d1, aba65a2f5df012a7f434e275f03d8cbd63c6fb56fc12a68d1dfcf877ae4bc7a3, additional_117, 8a88606731bd34583429af7deac06aa5de993adb02d231fad354ac7ad58e91dc, 7bca6b86363d9c8ab41964b3f2c90f9e0684e1e0c96e14e3617cefc6ebbf315f, 24bfbc52ddfe90eed7562e45f4764630f0befcd1825d6e995f459dbe3b59f23e, additional_138, additional_64, additional_116, 3d84df6267a2e73cd70576d10a8b6855250044ab8eea8f382060faeec4a09713, additional_125, 198f16c09501e24c2d3c71612c23c022fbb5d428f882da4e59e303fc14c4ac21, 9ad0369d0751e8632f8573904919702a44bfe10160cf77e2d2873c85a7714abb} &38\\ \hline
 Discourse& \citep{additional_55, additional_74, f980b6bb7db7c57e5d4c640e3cf54117d8a2e65667318ee8a2ac5e9320b38b86, additional_132, d039addf032f607f20bb25797f99cfba354c538edb6546f3e2d37c4bdf94e6a8, additional_97} &6\\ \hline
    \end{tabular}
\end{table}

\section{Analysis of the linguistic competence of PLMs}
\label{analysis_ling_competence}

In this section, we present a series of conclusions on the linguistic capabilities of PLMs, distilled from the different papers found across our survey. This information will be explained along several major linguistic levels: Syntax, (Lexico) Semantics, Morphology and Discourse.\footnote{While there is a study in our survey~\citep{additional_145} that touches pragmatics, we have not included a dedicated section to this level, because this paper does not study that linguistic level exclusively and pragmatics is not really their main focus.}
For each level, in order to organize the presented information, we will answer the following order of questions regarding the linguistic competence of PLMs:

\begin{enumerate}
    \item Is the studied type of linguistic information well-represented in PLMs? 
    \item Potential layerwise location(s) of that linguistic information --whether of the linguistic discipline in general or that of specific phenomena.
    \item Specific Transformer elements encoding that information --e.g. neurons, attention heads, etc.
    \item Geometry of contextualized representations --and whether a linguistic phenomenon of any type is somehow attested within the geometry of contextual embeddings.
    \item Overall conclusions --distilled from the points presented above.
\end{enumerate}

The reports presented in each section should not be taken as conclusive proof of a potential universal layerwise location of some specific linguistic phenomena, but rather as common occurrences happening specifically across several of the presented models that are trained in some specific languages.

\subsection{Syntax}
\label{syntax_conclusions}

The analysis of syntax in PLMs is the most studied linguistic discipline in the interpretability of these models, perhaps owing to the relative easiness of detecting and representing hierarchical structures within the embedding vector spaces of these representations --as initially demonstrated by~\citet{39e608673194e619191dbda4a5f348a8af0dca9b6603e901c28598351c10afb5}.

\subsubsection{Is syntactic information well represented in PLMs?}

Many works across our survey positively report on syntactic information being encoded in PLMs~\citep{additional_08, 8a88606731bd34583429af7deac06aa5de993adb02d231fad354ac7ad58e91dc} --although some works also note `unstable' probing results across different monolingual BERT models~\citep{a8c3c25f32aa64fd200681812f190528981a14639d8a2099497d7a1217c3cede}. 
Syntactic information is said to be more strongly encoded in these models compared to semantics~\citep{additional_112, 5cabfa60fe1197579bf66eb101e3d6110f007839095bc942706e6cdaf51eb843}, while it is also reported to have been acquired already during the pre-training of the models~\citep{e6c453545e421dd0d4a75b0a7b418660449471b7277016e71d042b71416aeae2, f79f09229def9cf277ff743fdb8c86896e0d5f9b3206423953cd48adeb6c809b} --and potentially being learned by the models first, during the earlier stages of their training, preceding that of semantic knowledge~\citep{e6c453545e421dd0d4a75b0a7b418660449471b7277016e71d042b71416aeae2}. 
For the linguistic phenomenon of pronominal anaphora, this seems to be well encoded in a specifically studied model: Transformer-XL~\citep{2f0b8010aeb5143941f96f1f398fc123f41f59674b6bc65229d61b61be8d3171}.
For multilingual models there is evidence of syntactic knowledge being transferred in mBERT across languages~\citep{6c0bcfd8ee5a86207dcb0ac639539437b705ee2b36ce47cc5e645d9d6a521200}, 
although sensitivity to syntactic knowledge in mBERT and mBART is also said to be different depending on either the language they are trained on or their used pre-training objectives~\citep{additional_122, ab5e694285556e053068e170ef7865a4a3997f4133b60adb6410048776d74eef}.

There also some works reporting negative or inconclusive results of the  internal encoding in PLMs of syntactic information. For instance, some works claim that word order is partially responsible for the apparent encoding of syntax-like structures in Transformer-based models~\citep{099659e8eec03507a3c4e46c25fb3566eaf52ce9e4f995024d0ec341730da708, additional_156}. In the same line, it has been shown that it is possible to pre-train PLMs on a corpora of texts with shuffled word order and still obtain good end results with the trained model~\citep{additional_147}, thus concluding that the statistical co-occurrence of words might be more important for the final model's performance. 
Conversely, other works find that Transformer-based models do not rely on positional information to derive syntactic trees~\citep{ab5e694285556e053068e170ef7865a4a3997f4133b60adb6410048776d74eef}, and that these models process positional information in lower layers but later change to a more hierarchical-based encoding in later layers~\citep{additional_07}.

Other linguistic phenomena were studied for which the authors did not find conclusive evidence of their encoding in the model, for instance: subject-verb agreement for less frequently-seen verb forms~\citep{additional_129}, reflexive anaphora~\citep{additional_07}, or implicit causality (IC) verbs~\citep{additional_74}. It was also shown that Chinese BERT shows degraded performance in attention heads when there is an increased distance between a
dependent and a head word~\citep{dee4ee9db0e8ad44bfb13b1493091a6206e87cda9327ec9a7900a5ae1bd2e9d7}.\footnote{This same work, however, as seen in section~\ref{syntax-transformer-elements}, gives otherwise support for the hypothesis of specific attention heads seemingly encoding syntactic phenomena in this language.}

Finally, other reported observation is that semantic and syntactic information may be conflated in PLMs, without a clear separation between the two~\citep{additional_107}. There is contradicting evidence, however, of the opposite idea~\citep{additional_160}, so no clear conclusions can be derived from these observations.

\subsubsection{Potential layerwise location of syntactic information in PLMs}

On a general basis, syntactic knowledge is stated to be located in either intermediate layers in BERT~\citep{additional_26, additional_29, 2e34ff32b63e31f6c144010143a0a51be955cb64583653296dfa868d529e6b1d, twentyfour_04} or alternatively in middle to upper layers~\citep{additional_159,additional_165,bertology}.\footnote{\citet{2e34ff32b63e31f6c144010143a0a51be955cb64583653296dfa868d529e6b1d} find that while syntax trees are embedded in intermediate layers, this does not automatically entail that the model may be using this information for downstream tasks.} 
In a multilingual setting, mBERT ranks best in its 6th layer (of a total of 12) for morphosyntactic information encoded across a series of typologically-different languages\footnote{These include Afrikaans, Arabic, Chinese, Croatian, Finnish, Hebrew, Korean, Marathi, Slovenian, Spanish, Tagalog, Turkish and Yoruba.}~\citep{additional_117} --or, alternatively, layers 7 and 8 of that model~\citep{additional_58}.
On a related note, mBERT and XLM-R seem to be capable of detecting syntactic anomalies in intermediate layers~\citep{3d84df6267a2e73cd70576d10a8b6855250044ab8eea8f382060faeec4a09713}. Exceptionally, mBART models --primarily testing the Russian language, but complementing their analysis in English as well-- are found to encode syntax better in upper layers: specifically, the 11th --penultimate-- and the 12th --last-- layers~\citep{additional_142}. In this same work, however, other tested models, such as mBERT and XLM-R, display similar layerwise tendences to other studies. The different results presented in the case of mBART could be attributed to the different pre-training objective used for this model, which may alter its layerwise location.\footnote{This is a fact that is explained by~\citet{41587c8da1c20c412c07ea2233564dde0a8272a2e1b3515f0e745824e9f9e59a}.}

There is another study,~\citet{additional_150}, that studies constituency grammar in attention heads weight matrices, and report better results for BERT in upper layers and for RoBERTa in middle layers. However, they also acknowledge that both models do ``not fully learn much constituency grammar knowledge''.
There are some observations as well on BERT potentially imitating the distribution of tasks in a classical NLP pipeline across its layers~\citep{7d129a1596d1adb62bc05a641d18f71f70d840c7faee2ab9a72f8c8ea7a9c8f1} --including syntactic-based tasks. This account, however, is disputed by other work~\citep{6bde56f74fa302633de19099fc56d2a3247291067a76b5473c027460d6d2eac5}.

Opposed to the above statements, syntax has also been found to be better represented instead in lower layers of BERT~\citep{198f16c09501e24c2d3c71612c23c022fbb5d428f882da4e59e303fc14c4ac21,eef9d3252ff314123f8ef485036e676bd933b3a0d0647e23992e5522eec1b178, additional_138}, but their provided conclusions may be hindered by several factors. On the one hand, \citet{198f16c09501e24c2d3c71612c23c022fbb5d428f882da4e59e303fc14c4ac21} warn that their probing methodology may fail to reveal linguistic knowledge in the middle layers, and ~\citet{eef9d3252ff314123f8ef485036e676bd933b3a0d0647e23992e5522eec1b178} account for a different architecture consisting of an encoder in a machine translation system.\footnote{In their case, they observe that it is the first three layers --out of a total of six-- that encode most syntactic information, and that data on sentence length starts to disappear starting from the third layer. We have to take into account, however, that this encoder is part of an encoder-decoder NMT pair, and that its internal working is likely geared towards the decoder module, unlike other models which are encoder or decoder-based only.} Finally, although~\citet{additional_138} demonstrate the existence of several dependency parse-like phenomena being encoded in early layers of BERT --and not being tied to any particular neurons--, they do not probe all layers from the model but only a selection of them: 1, 4, 8 and 12. On the other hand,~\citet{additional_138} simply demonstrate that earlier layers of BERT solve syntax probing tasks with lower-dimensional subspaces --one of their hypothesis, used as their basis for their probing method-- compared to the rest of the layers, something that does not necessarily imply that syntactic information may not be encoded in ongoing layers as well.

There is another hypothesis, however, that gives lower layers a potential joint role in the encoding of syntax alongside middle layers~\citep{a8c3c25f32aa64fd200681812f190528981a14639d8a2099497d7a1217c3cede} --with lower layers being where most syntactic information ``either emerges or becomes accessible'' and middle layers where ``the most overall information is located''~\citep{a8c3c25f32aa64fd200681812f190528981a14639d8a2099497d7a1217c3cede}.\footnote{On a related note, although \citet{198f16c09501e24c2d3c71612c23c022fbb5d428f882da4e59e303fc14c4ac21} show that syntactic information is mostly found in lower layers, this work also finds that specific syntactic phenomena --e.g. closed-class words against open-class ones-- are provided by some layers of BERT spread throughout the model.} 
It has also been found that while Transformer-based models mainly process positional information about tokens in lower layers, they later seem to switch to a more hierarchical-based encoding in their higher layers~\citep{additional_07}.

These aforementioned reports mainly focus on monolingual models in the English language, \textbf{but we should note that a layerwise location could be different for other languages.} In one specific instance --subject/verb agreement in Italian models-- we have identified a commonality with English-based models: this information is more strongly encoded in central to upper layers~\citep{788f3f1df1a8c3daa1e4bbfdf6afbe350660000ad607b64a238cd4946bbc63f5}.
However, it has also been stated that for models trained in typologically different languages --e.g. English, Korean and Russian-- the layerwise location of a series of morphosyntactic phenomena can change depending on the language~\citep{9517593e2fca883a1391c8d8068f4426fa7663cab61074fe7e784297edd020b7}.\footnote{This work, for instance, observes that English and Korean models obtain good probing results with overall less layers involved, whereas Russian requires far more layers.}
Similarly, syntax probing results across monolingual models in different languages have proven unstable~\citep{a8c3c25f32aa64fd200681812f190528981a14639d8a2099497d7a1217c3cede}.
In Table~\ref{tab:layerwise_syntax_location} we have included instances of some additional specific syntactic phenomena encoded in a series of layerwise locations.

\begin{table}
\caption{Potential layerwise location of several specific syntactic phenomena in PLMs}
    \label{tab:layerwise_syntax_location}
    \small
    \centering

    \begin{tabular}{>{\centering\arraybackslash}p{10cm}cl}
    \hline
         \textbf{Phenomenon} & \textbf{Model} & \textbf{Location (layer)} \\ \hline
         Order of the subject/object with respect to the verb~(\citet{additional_64})&  BERT&Middle layers\\ \hline
 Subordination and verbal predicate structure~(\citet{additional_64})& BERT&Middle layers\\ \hline
 Attention heads that induce constituency grammar~(\citet{additional_150})& BERT&Upper layers\\ %\cline{2-3}
  & RoBERTa&Middle layers\\ \hline
 Syntactic anomalies~(\citet{3d84df6267a2e73cd70576d10a8b6855250044ab8eea8f382060faeec4a09713})& mBERT, XLM-R&Middle layers\\ \hline
  Subject-verb agreement and null-subject~(\citet{788f3f1df1a8c3daa1e4bbfdf6afbe350660000ad607b64a238cd4946bbc63f5})& Italian BERT&Middle layers\\ \hline
    \end{tabular}
\end{table}

Other works, on the other hand, have questioned the overall notion of layer-localized syntactic knowledge. 
In this sense, a hypothesis has been put forward for the existence of syntax-specific neurons that are spread throughout the entire model~\citep{additional_90}.\footnote{The latter applies for the case of BERT, although these syntax-specific neurons are said to be localized in the final layer in other architectures such as ELMO (non-Transformer) or XLNet~\citep{additional_90}.} At the same time, however, other observations counterclaim that syntactic phenomena may not be tied to any specific neurons~\citep{additional_138}.\footnote{This same work, however, argues that the studied syntactic phenomena are encoded in low-dimensional subspaces in lower layers of BERT.}
Another hypothesis presents the idea of a different internal organization that seems to relate to a linguistically-based generalization~\citep{6bde56f74fa302633de19099fc56d2a3247291067a76b5473c027460d6d2eac5}.\footnote{This is the same work that opposes the statement, presented by~\citet{7d129a1596d1adb62bc05a641d18f71f70d840c7faee2ab9a72f8c8ea7a9c8f1}, that BERT-based models may imitate the distribution of a classical NLP pipeline.}

\subsubsection{Specific Transformer elements encoding syntactic phenomena}
\label{syntax-transformer-elements}

Some \textbf{attention heads} seem to be specialized in specific dependency relations and syntactic phenomena~\citep{4ae9f84833b4deb3ea3aba5a62bfab93b0d6c27c440be7f2be4ec98772c41f0a, dee4ee9db0e8ad44bfb13b1493091a6206e87cda9327ec9a7900a5ae1bd2e9d7}, with some heads potentially acting as proxies for constituency grammar~\citep{additional_150} --reportedly found in the higher layers for BERT and in the middle layers for RoBERTa.
In the case of BERT models trained for Chinese, these have been shown to feature some attention heads that have learned specific dependency relations and syntactic phenomena in that language~\citep{dee4ee9db0e8ad44bfb13b1493091a6206e87cda9327ec9a7900a5ae1bd2e9d7} --with some hidden states also showing ``some competence in encoding syntactic knowledge''.
However, syntactic information is likely not uniformly distributed across attention heads: it has been shown that a single specific dependency relation can be spread across multiple heads or, conversely, an attention head may encompass several dependency relations at the same time~\citep{additional_81}. 
In a similar line, for the specific case of a Machine Translation encoder, there are reports of at least one attention head being present per layer in that model that encodes many syntactic relations~\citep{eef9d3252ff314123f8ef485036e676bd933b3a0d0647e23992e5522eec1b178}.\footnote{This study analyzes the attention weights in the heads of each layer via a probing task that induces parse trees, then comparing them against those of a gold English-based treebank. However, this work provides no specific mentions of which syntactic phenomena they studied and rely instead on extracting unlabeled parse trees in each studied language pair that has English as its source. They then provide the results obtained per attention head and layer.}
In the context of a similar model for Machine Translation as well, a set of heatmaps extracted from specific attention heads from an encoder have been found to contain a series of graphical patterns --in the form of `balustrades', similar to stairs-- that seem to roughly correspond to syntactic phrases that are being processed from input sentences~\citep{additional_09}. Furthermore, this work compares constituency trees automatically built from these patterns to those produced by a syntactic parser, to favorable results. However, their evaluation strategy --in which they do not account for alternative syntactic structures in heatmaps-- limit potential linguistic claims about this discovery.

Regarding the case of \textbf{neurons}, for mBERT, it has been found that the same sets of neurons in this model seem to encode several morphosyntactic phenomena across languages~\citep{7bca6b86363d9c8ab41964b3f2c90f9e0684e1e0c96e14e3617cefc6ebbf315f} --including Arabic, English, Finnish, Polish, Portuguese and Russian.
However, for the phenomenon of subject-verb agreement, although there are reports of `significant' neuron overlap~\citep{additional_171} in autoregressive multilingual language models across several typologically different languages --English, French, German, Dutch and Finnish--, the same is not reported for multilingual masked language models --e.g. mBERT. This work also finds ``two distinct layerwise effect patterns and two distinct sets of neurons used for syntactic agreement, depending on whether the subject and verb are separated by other tokens''.

\subsubsection{Geometry of contextualized representations}

Contextual embeddings are found to encode both syntactic and semantic information, albeit in separate linear subspaces~\citep{867bde36723c79dc106d857c379bcdf584fb5f04a18b34d034ff3da4133d59f7}. Additionally, the linear subspaces where semantic information is encoded seem to be low-dimensional~\citep{867bde36723c79dc106d857c379bcdf584fb5f04a18b34d034ff3da4133d59f7}.
There are also distinct subspaces, located across all the layers of BERT, that separately encode linguistic hypernymy, dependency syntax and word position~\citep{additional_87}. 
For agreement, albeit focusing on PoS and dependency syntax,~\citet{additional_138} mention that linguistic variables are encoded in low-dimensional spaces --and not tied to any particular neurons-- and prove, through some ablation experiments, that BERT is found to rely ``on subspaces with as few as 3 dimensions to make fine-grained part of speech distinctions when enforcing subject–verb agreement''. These subspaces also display a certain sense of hierarchy.
For mBERT, there are reports of a shared syntactic subspace, with layers 7 and 8 of that model showing the best results~\citep{additional_58}. However, this same work also noted that a small number of their observed results in those layers may have been caused by word order-related phenomena, and not linguistically-based reasons.
It has also been shown, however, that different linguistic concepts --including syntactic ones-- may tend to overlap in the latent space within contextual embeddings ``to a varying degree''~\citep{additional_165}.

Finally, referring to information contained within specific tokens, \citet{1bce61f930980f7a779b24aeb82bd4deb4cadfeb316b5253fd3793dda36e29eb} find that, for morphosyntactic information in BERT, ``the most informative word representation is the one that correspond to the last token of each input sequence and not [...] to the [CLS] special token''.

\subsubsection{Overall conclusions on syntactic competence of PLMs}

Overall, although there are some specific syntactic phenomena in which the studied models are found to fail systematically --such as agreement errors in low-frequency verb forms, a weak encoding of reflexive anaphora or an excessive reliance of these models on word order--, syntactic information seems to be overall well represented in a series of studied PLMs.

A potential layerwise location of this information is still disputed, with a dichotomy between the notion of middle layers being the most prominent in this process, compared to the combination of lower and intermediate layers. Nevertheless, most of the provided conclusions seem to point at middle layers of BERT-based models in the English language to be of great relevance to the encoding of syntactic information, with some degree of intervention of the lower layers to process this information as well. 
The pre-training objective used to train a model has also been shown to alter syntactic layerwise location, as is the case with mBART models.
Contrary to any layerwise accounts, other work points to syntactic information being spread throughout all layers of a model. As such, no definitive conclusions can be provided on the layer location of syntactic information, although some recurring patterns seem to appear across models belonging to the same family and trained in the same language.

There is also evidence of specific attention heads encoding syntactic phenomena, although other works point to this information being sparsely spread across many attention heads found in different locations of a model. In the case of neurons, for multilingual models, although there are reports of shared sets of neurons being reused for the same linguistic phenomena across different languages in these models, there are some specific phenomena --e.g. agreement-- for which this statement does not hold, depending on the multilingual model in question: those neurons do seem to exist in autoregressive multilingual models, but not in masked ones.

The geometry of contextualized representations seem to encode both syntactic and semantic information in separate linear subspaces, and there are other subspaces as well that encode dependency syntax and word position also separately. Despite these reports, there is contradicting evidence of other syntactic concepts not being so separable. As such, no clear conclusions can be derived from these overall observations.

\subsection{(Lexico) Semantics}

We have found a large number of studies on semantics in our survey, which proves a general interest in the research community to understand the extent over which Transformer-based contextual language models seem to be able to encode the knowledge of the meaning of words and texts. Some of the studies in this area also investigate on lexicosemantic phenomena.

\subsubsection{Is semantic information well represented in PLMs?}
\label{conclusions_specific_semantic_phenomena}

Aside from syntactic knowledge, semantic knowledge seems to be overall well represented in state-of-the-art pre-trained PLMs~\citep{7d129a1596d1adb62bc05a641d18f71f70d840c7faee2ab9a72f8c8ea7a9c8f1, additional_08, additional_177}, with BERT representations ``satisfy[ing] two desiderata for psychologically valid semantic representations: i) they have a stable semantic core which allows people to interpret words in isolation and prevents words to be used arbitrarily and ii) they interact with sentence context in systematic ways, with representations shifting as a function of their semantic core and the context''~\citep{7d129a1596d1adb62bc05a641d18f71f70d840c7faee2ab9a72f8c8ea7a9c8f1}.
BERT is also found to ``split core semantic roles into many fine-grained categories, and seem[s] to encode broad notions of syntactic and semantic structure''~\citep{additional_92}.

During the pre-training of the models, syntactic capabilities are rapidly acquired while semantic knowledge is learned in later stages of the model's training, in a progressive manner~\citep{e6c453545e421dd0d4a75b0a7b418660449471b7277016e71d042b71416aeae2}.
In the case of RoBERTa, it has been shown that the model is slower at learning facts and commonsense knowledge, depending as well on the domain~\citep{additional_124}.

Regarding more specific semantic phenomena, we can also provide the following conclusions:

\begin{itemize}

\item BERT encodes polysemy well~\citep{ff69f49e3a5e82a147575bc0dfcc94bcf3848fc9c90595829952ec3bbb706041}, across several languages --English, French, Spanish and Greek. 
However, polysemous representations are found to be better in monolingual BERT models compared to mBERT --a fact that is partly blamed on the use of an English-oriented tokenizer for the latter, which skews multilingual representations to that language~\citep{ff69f49e3a5e82a147575bc0dfcc94bcf3848fc9c90595829952ec3bbb706041}. 

\item BERT seems to possess knowledge of dates, scalar and measurable values, and is capable of distinguishing between small and large numbers~\citep{additional_92}.

\item The modeling of verb-argument structure --relying on the Dowty theory of thematic proto-roles~\citep{dowty91}-- is well encoded in BERT~\citep{4809831c190d9712231d7022e8d438c0e1b0cb666c169632fdc8c5f787e8d828}. 

\item In multilingual models, the encoding of `subjecthood' --i.e. the notion of subject in a sentence-- is not only based on syntactical factors but also dependent on semantic ones~\citep{additional_125}.
Further, multilingual models seem to capture lexical features well --although not so on nominal and verbal features~\citep{41128f3f273ba991eaff22672cb2714ea9e471d273d3c6f4dcfa590d5e173fdc}.\footnote{These features correspond to a series of typological language features described by two datasets used by the experiments in this work: the `World Atlas of Language Structures' or WALS~\citep{WALS}, and the ` Syntactic Structures of the World's Languages' or SSWL, findable at~\url{https://terraling.com/groups/7}.}

\end{itemize}

Another interesting area is that of idiomatic expressions: 
BERT is able to distinguish between the literal and figurative meanings of idiomatic expressions --also known as PIEs--, and also seem to encode their idiomatic meaning as well~\citep{730b1312face40c86e1bb39615e5032b054829cbf9d47753d186adac061a46ff}.

Outside of traditional semantic notions, Transformer-based models also perform well on NLP tasks that require semantic knowledge --e.g. word sense disambiguation (WSD)~\citep{c150031d36ba8c897fa8cc81b743700a854c824c42ae375bdb6baaa7365c6a74} or entity matching (EM)~\citep{4a343f0a56c9dea29ff8484c457c3a894f5b8d37c85f407b2f7723c96ccc3494}. For the latter, the authors observed that their studied model, BERT, recognizes ``the structure of EM datasets and extracts from the entity descriptions semantic knowledge that goes beyond the pair-wise association between tokens''. 
\citet{d0eb7251a950c57432892c84c25caa8e8a936348d985236d352da233dfdd18f8} also report good results for a cross-domain noun WSD task, but they admit at the same time that their analyzed models may not be as performant when exposed to real-world WSD datasets with artifacts and related issues.

There also some works reporting negative or inconclusive results. 
We first refer to an ongoing hypothesis, which we also discuss in section~\ref{syntax_conclusions}, that claims that \textbf{syntax seems to be more strongly encoded in PLMs compared to semantics}~\citep{additional_112, 5cabfa60fe1197579bf66eb101e3d6110f007839095bc942706e6cdaf51eb843}.
In this line,~\citet{0ea4eaa575b9fe22c52fe5eccb54c7449f1f24c2652c3d89ae8e4b96eb5d6720} report excellent syntactic performance in their studied model SpanBERT, but they observe that it does not capture domain-specific semantic concepts. They analyze scientific documents, and find that ``its semantic understanding of scientific domain documents is weak which further leads to cascading problems for the coreference resolution task''.
Factual reasoning has been found to be based on context rather than abstraction or composition in both BERT and RoBERTa~\citep{b8243b9e6ab5a5f1d7f0777ea2928465082d00215d5ca57bc58a16d68efca413}. Reasoning abilities have also been shown to not be stably acquired in RoBERTa~\citep{additional_124}, and BERT has also been shown to not rely on frame semantics~\citep{additional_21}.

For the phenomenon of negation, which has been widely reported as not being well represented by PLMs --see~\citet{bertology} for an examination on the topic--, we have found some mixed results. While some studies report it is well encoded in their studied models~\citep{8a88606731bd34583429af7deac06aa5de993adb02d231fad354ac7ad58e91dc}, others provide different conclusions~\citep{9741c619a3f09f7a316e0907909293b33562493c6ed4f5c033078d710a614ec7, 6d5cc37605e9d476ec1fe06906f83e72c4ca0fc45caaea21ecae1aebbcd6c50e}. For instance, probing results for negation scope detection point this information to be located in specific layers in BERT-BASE and RoBERTa-BASE, but the results are not so conclusive for the LARGE variant of both models~\citep{9741c619a3f09f7a316e0907909293b33562493c6ed4f5c033078d710a614ec7} --which might be encoding negation in another way.
Other work~\citep{6d5cc37605e9d476ec1fe06906f83e72c4ca0fc45caaea21ecae1aebbcd6c50e} reports that, when testing models on pairs of sentences containing positive and negative affirmations (e.g. `The boy played the piano' against `The boy \textit{did not} play the piano'), the contextual embeddings for the individual words in the negative sentences showed a different internal representation compared to those of positives ones --whereas the authors had been expecting not to be there much of a difference. They also observed the same tendency when analyzing sentences that had been transformed to a passive voice.

\subsubsection{Potential layerwise location of semantic information in PLMs}

Several works across our survey have attempted to locate general semantic information or phenomena in specific layers of a model. 
Overall, there does not seem to be a consensus: some works report middle layers~\citep{82436c5c586808f7e49bec663f288b6637ecd907f5086aac3132cb16c3546545} for its potential location or middle to higher layers~\citep{b7ec8a05b964b2a7bf7b0016b96e3c83867e03b84978fbe4356572bde8a3f561, twentyfour_04} --and higher layers for the specific case of an encoder in a Transformer-based Machine Translation encoder-decoder system~\citep{eef9d3252ff314123f8ef485036e676bd933b3a0d0647e23992e5522eec1b178}. Conversely, another work has also found semantic information to be better found in lower layers~\citep{867bde36723c79dc106d857c379bcdf584fb5f04a18b34d034ff3da4133d59f7} and, for a specific model, LABSE, tested in English and Russian datasets, semantic information has been found to be encoded in lower to middle layers~\citep{additional_142}. The latter's layerwise location of semantic information, however, may have been determined by its different pre-training objectives.\footnote{This is in line with a remark by~\citet{41587c8da1c20c412c07ea2233564dde0a8272a2e1b3515f0e745824e9f9e59a} that prove that the pre-training objective or underlying architecture of a Transformer-based model can lead to linguistic knowledge being located in different parts depending on a studied model.} Interestingly, for the case of~\citet{867bde36723c79dc106d857c379bcdf584fb5f04a18b34d034ff3da4133d59f7}, this work argues that semantic information is encoded in low-dimensional subspaces in lower layers, a conclusion that is strikingly similar to that of~\citet{additional_138} --which claim the same, but for syntactic information, and is also a work that presents lower layers as well regarding the layerwise location of syntactic phenomena.
On the other hand, there seems to be a consensus on lower layers of Transformer-based models to be mostly lexical in nature by several works across our survey~\citep{additional_156, additional_165, additional_61}.

\begin{table}
\caption{Potential layerwise location of some specific semantic phenomena in a series of PLMs}
    \label{tab:layerwise_semantic_location}
    \small
    \centering
    \begin{tabular}{>{\centering\arraybackslash}p{9.2cm}>{\centering\arraybackslash}p{1.4cm}>{\raggedright\arraybackslash}p{2.6cm}}
    \hline
         \textbf{Phenomenon} & \textbf{Model} & \textbf{Location (layer)} \\ \hline
         Metaphors~(\citet{additional_176})&  BERT&Middle layers\\ \hline
 Semantic similarity~(\citet{755a82b7a6fbf89197a027287f273eb7f4f88301aa62fc4510f6350c6989590b})& BERT&7th layer (out of 12)\\ \hline
 Compounds~(\citet{additional_189})& BERT-Base&Layer 8/9 (out of 12)\\
 & BERT-Large& Layer 19/20 (out of 24)\\ \hline
 Hyperboles~(\citet{77b165db81953eb8217e121eb7ae050919a23cab7146ee76c4a975fb9f0072b7})& BERT&Upper layers\\ \hline
  Relatedness~(\citet{755a82b7a6fbf89197a027287f273eb7f4f88301aa62fc4510f6350c6989590b})& BERT&12th layer (out of 12)\\ \hline
 Causativity and non-causativity of events denoted by a verb~(\citet{2263597f16afdc0b04e5c64cecd337e8f2728c07a6a3ee46b06502245e7e5857})& BERT&Upper layers\\ \hline
    \end{tabular}
\end{table}

A related hypothesis, also presented in section~\ref{syntax_conclusions}, is that PLMs may mimick the classical NLP pipeline across their layers --including semantic-oriented tasks such as NER, semantic roles and coreference~\citep{additional_08}. This idea, however, has also been questioned~\citep{6bde56f74fa302633de19099fc56d2a3247291067a76b5473c027460d6d2eac5}.
In Table~\ref{tab:layerwise_semantic_location} we report on some specific semantic phenomena being found in specific layers of BERT models across several studies.

Contrary to any layer-localized reports, other works do not support the idea of semantics being localized in specific layers of a model~\citep{6bde56f74fa302633de19099fc56d2a3247291067a76b5473c027460d6d2eac5, additional_90, 198f16c09501e24c2d3c71612c23c022fbb5d428f882da4e59e303fc14c4ac21}. In this sense, for instance, observations have been made on the potential existence of specialized neurons that process semantic phenomena, but which are found distributed throughout the entire model rather than localized in specific layers~\citep{additional_90}\footnote{Except for XLNet, where this information is processed in specific neurons that are mostly found in lower layers of the model~\citep{additional_90}.}, or that semantic phenomena are seemingly spread across all layers of a model~\citep{198f16c09501e24c2d3c71612c23c022fbb5d428f882da4e59e303fc14c4ac21}. Other work has pointed in the direction of an internal organization that does not resemble a classical NLP pipeline but seems to contain a linguistically-motivated internal organization~\citep{additional_90}, presenting a custom layerwise distribution for the processing of this type of information.

\subsubsection{Specific Transformer elements encoding semantic information}

We have found a series of reports on specific \textbf{neurons} seemingly encoding information about several semantic phenomena:

\begin{itemize}

\item Specific neurons are able to capture the difference between arguments and adjuncts in PLMs~\citep{c4b9ab20f2a2bf267e75987b9d9fa6ac2fe15ed0b9b90c0b65922c85136532b4} --although this work reports better results in a fine-tuned model compared to a pre-trained one.

\item Closed-class words, such as interjections, are found to be handled ``using fewer neurons compared to polysemous words (such as nouns and adjectives)''~\citep{additional_90}.\footnote{Additionally, other work~\citep{198f16c09501e24c2d3c71612c23c022fbb5d428f882da4e59e303fc14c4ac21} point at lower layers of a model handling closed-class words --with some exceptions, such as numbers or personal pronouns-- whereas higher layers do the same for open-class words.}

\end{itemize}

Regarding \textbf{attention heads}, BERT has been found to be overparametrized, with some attention heads being redundant and with the user being able to be prune them without affecting model performance in semantic tasks~\citep{additional_21}.

\subsubsection{Geometry of contextualized representations}

The geometry of contextual vectors supports the idea of the existence of different subspaces that encode syntactic and semantic information~\citep{867bde36723c79dc106d857c379bcdf584fb5f04a18b34d034ff3da4133d59f7}
--with a specific subspace, located across most layers of BERT, that encodes linguistic hypernymy separately from other spaces that encode dependency syntax and word position~\citep{additional_87}.
The abstract semantic notion of plausability is claimed to be ``one of the organizing dimensions of the underlying distributional spaces for middle and late layers''~\citep{a58e1f06742805ce6fa99fc3bd0db86c2fc39b5c7404bea798342dc91285226f}. 

\subsubsection{Overall conclusions on (lexico) semantic performance of PLMs}

In conclusion, there are many positive assessments of semantics being well encoded in Transformer-based PLMs, including numerous reports in specific phenomena such as polysemy, scalar and measurable values, `subjecthood' in sentences, etc. However, at the same time, similarly to syntax, a series of studied models are found to possess as well some systematic failings in processing some other specific phenomena --e.g. negation, words of high functionality, comparative correlatives, the phenomenon of non-composionality, etc. These errors are further confounded by an ongoing hypothesis that proves that the encoding of semantic information in PLMs seems to be weaker compared to syntactic information. We should not forget, however, that we also presented other evidence as well of other semantic phenomena which were well encoded by these models, so we should not quickly dismiss the semantic capabilities of these models --we should simply acknowledge the existence of some limitations alongside several strengths.

Regarding the layerwise location of semantic information in PLMs, this is highly disputed among works, with its supposed location depending on the studied model architecture. Although a majority of works vouch for middle layers or middle to higher layers in BERT-based models, there is another work that finds lower layers more favorable instead. There seems to be a consensus, however, on lower layers being primarily involved in the processing of lexical information in BERT. As with syntax, however, other works dismiss a layerwise hypothesis altogether and point instead to semantic-specialized neurons that are spread throughout the entire models. As a result, no general conclusions on the layerwise location of semantic information on PLMs can be deduced from the provided conclusions.

We have not been able to find many conclusions from our body of works on specific neurons encoding semantic phenomena, except for two interesting remarks: there are some specific neurons that are able to distinguish arguments from adjuncts, and closed-class words are apparently handled using fewer neurons in these models compared to polysemous words. We can also provide very few information on specialized attention heads in semantic information, outside of reporting that they redundantly encode this type of information in BERT models and thus a number of them can be pruned without major issues.

For the geometry of contextualized embeddings, there seem to be different subspaces within these representations that separately encode syntactic and semantic information, with one subspace apparently encoding linguistic hypernymy.

\subsection{Morphology}

Almost all works that study morphology in our survey do not analyze this linguistic level on an exclusive basis, but also combine it with the study of either syntax or lexico-semantics. Only two research papers~\citep{bb77fe70885ba02cbc438e9ba43185fcf11a46359ee71738a2e39975ee07fc30, fd02effe5f3f2424aee1e7b6f4be2bdf4619419336b3b32077a706955b79ca81} study morphology per se, whereas 19 research papers study syntax and morphology, and 15 other works do so for syntax, morphology and lexico-semantics. Only a single paper~\citep{2f0b8010aeb5143941f96f1f398fc123f41f59674b6bc65229d61b61be8d3171} studies morphology and lexico-semantics combined.
We should always note that the study of morphology in PLMs will be hindered in many cases by the underlying tokenizer used by each model --wordpiece for BERT, BPE for RoBERTa and GPTs, etc.--, which divides texts into individual, arbitrary tokens that are determined not by their morphological content but derived from text compression techniques. As such, a same morpheme will in most models be expressed by different tokens in a model's vocabulary depending on its surrounding text, complicating its study.

\subsubsection{Is morphological information well represented in PLMs?}

Due to the reduced number of studies in this area, we have not found any major claim on positive performance of Transformer-based models on morphological information as a whole. Reports on more specific morphological phenomena, however, representing instances of positive, as well as negative, occurrences of morphological performance across many PLMs are described in the following sections.

We have not been able to find many works that provide negative or mixed opinions on the general morphological competence of PLMs. However, we can refer to a work, presented by~\citet{aba65a2f5df012a7f434e275f03d8cbd63c6fb56fc12a68d1dfcf877ae4bc7a3}, which we also discussed in section~\ref{syntax_conclusions}, that analyzes morphological phenomena such as grammatical number or tense, in English and Polish models. They find that their probes trained on BERT-based models perform worse --although still showing good scores-- compared to non-Transformer-based models. This research work, however, only probes the penultimate layer of the BERT model.

\subsubsection{Potential layerwise location of morphological information in PLMs}

Syntax and morphology seem to share the same layerwise locations in PLMs:  middle to higher layers~\citep{additional_165}.
For mBERT, probing experiments \citep{additional_117} point to the 6th layer (out of a total of 12) of this model as the best performing for morphosyntactic information.
On the other hand, it has also been observed that the encoding of morphosyntactic properties --e.g. agreement, grammatical gender, etc.-- can change its layerwise location in monolingual models depending on the studied language~\citep{9517593e2fca883a1391c8d8068f4426fa7663cab61074fe7e784297edd020b7}.\footnote{Specifically, \citet{9517593e2fca883a1391c8d8068f4426fa7663cab61074fe7e784297edd020b7} study BERT-based models trained for English, Korean and Russian.} This is a fact that had been addressed as well for syntactic and lexico-semantic information.

Regarding more specific phenomena, we can also find the following observations:

\begin{itemize}

\item Number information is transferred from a noun to its head verb between BERT's 3rd and 8th layers~\citep{additional_162} --although most of this information is also passed indirectly through other tokens in a sentence.
\item For French, number information for the object-past participle agreement is locally distributed within the context tokens~\citep{2ab1cd858abb87043053d4aa1c3d8fcc9ab7381f1788ee5e77213ad54e32037d}. This work also states that ``if this information is encoded in a small amount of highly correlated dimensions, it is also fuzzily encoded in a redundant way in the remaining dimensions''. 

\item Regarding ungrammatical examples and linguistic anomalies, anomalous inputs are found to be out-of-domain in higher layers~\citep{additional_116}, with ``morphosyntactic anomalies [...] recognized as out-of-domain starting from lower layers compared to syntactic anomalies''.

\item For grammatical number, tense information, word-level and phrasal-level inversion, \citet{additional_108} conclude the following in relation to the higher layers of the models:

\begin{quote}
    "while most of the positional information is diminished through layers, sentence-ending tokens are partially responsible for carrying this knowledge to higher layers in the model. BERT tends to encode verb tense and noun number information in the \#\#s token and that it can clearly distinguish the two usages of the token by separating them into distinct subspaces in the higher layers [...]".
\end{quote}

\end{itemize}

\subsubsection{Specific Transformer elements encoding morphological information}

Several works support the idea of certain \textbf{neurons} being apparently specialized in the encoding of morphological information of different type~\citep{additional_90, 7bca6b86363d9c8ab41964b3f2c90f9e0684e1e0c96e14e3617cefc6ebbf315f}, with fewer neurons seemingly involved in the encoding of morphology compared to syntax --although these neurons are distributed along all layers in the BERT model~\citep{additional_90}.
In mBERT, morphosyntactic information seems to be encoded as well across languages by the same sets of neurons~\citep{7bca6b86363d9c8ab41964b3f2c90f9e0684e1e0c96e14e3617cefc6ebbf315f}, developing in the process a ``cross-lingually entangled notion of morphosyntax''.
Regarding word structure, several studies on Chinese models report specific \textbf{attention heads} that are specialized in this phenomenon~\citep{additional_75}.

\subsubsection{Geometry of contextualized representations}

BERT has been found to rely on a linear functional encoding of grammatical number to solve the number agreement task in English~\citep{additional_162} --and gender as well as number in the case of Spanish~\citep{additional_161}. \citet{additional_162} also provides evidence that nouns and verbs do not have a shared functional encoding of number, and that English BERT relies instead on disjoint sub-spaces to extract this information.

\subsubsection{Overall conclusions on morphological performance of PLMs}

A layerwise location of morphological information cannot be determined with ease and thus cannot be generalized. Although there are some reports of morphological information being shared with syntax in middle layers of BERT and mBERT, there are also some more specific morphological phenomena --e.g. number information, tense, morphosyntactic anomalies, etc.-- that are found in other disparate locations, spread throughout the entire models. It has also been shown that the language used for pre-training a model can change the layerwise location of morphological information.

There is evidence of specific neurons specialized in morphological information, spread throughout all layers in the case of BERT, and that there are fewer neurons processing morphology compared to syntax. In the case of mBERT, there are also sets of shared neurons jointly encoding morphological information from different languages. These overall conclusions on neurons, however, should be taken with care, since we have also presented, for other linguistic levels different to morphology --e.g. syntax or semantics--, evidence of both linguistic-specialized neurons as well as of other works that proved the opposite hypothesis under some circumstances. Finally, regarding attention heads, in several models trained for Chinese, some heads are seemingly specialized in encoding word structure in that language.

\subsection{Discourse}

Among the six works in our survey that study discourse in PLMs\footnote{Due to the scarcity of works in this topic, we are not following in this section the same division as in the other linguistic levels.}, three of them also study syntax simultaneously~\citep{additional_55, additional_74, f980b6bb7db7c57e5d4c640e3cf54117d8a2e65667318ee8a2ac5e9320b38b86}. The remaining three research papers~\citep{additional_132, d039addf032f607f20bb25797f99cfba354c538edb6546f3e2d37c4bdf94e6a8, additional_97} in that list analyze discourse exclusively. We will present some of the most relevant conclusions reached by several of these works in relation to discourse:

\begin{itemize}

\item \citet{additional_74}, studying implicit causality verbs, find that discourse structure only influences PLMs' behavior for reference, not syntax, despite model representations that encode the necessary discourse information.
\item \citet{f980b6bb7db7c57e5d4c640e3cf54117d8a2e65667318ee8a2ac5e9320b38b86} analyze the phenomenon of disfluency, by comparing pairs of fluent and disfluent sentences, and find that, the deeper the layer in the analyzed model, the less sensitive the obtained representations become to disfluency --i.e. the analyzed pairs of fluent and disfluent sentence embeddings become increasingly similar one to the other. They claim the attention mechanism may explain this phenomenon.
\item \citet{d039addf032f607f20bb25797f99cfba354c538edb6546f3e2d37c4bdf94e6a8} study coherence between clauses and discourse relations, although focusing on the use of abstracts from scientific papers --this is the reason why they use SciBERT aside from a regular domain BERT model. They find that both BERT and SciBERT, from their pre-training, seem to encode coherence ``to some extent'', but they also observe that these models do not do the same for the semantics of the discourse relations they studied. On the other hand, they claim that coherence links are captured in contextual embeddings, the same as for discourse relations --despite the models not encoding the semantics of the latter.
\item Finally, \citet{additional_97} study Rhetorical Structure Theory (RST)\footnote{Area of study within Discourse Structure Theories (DSTs). The latter is a discipline whose aim is to represent the discourse of a text in a structured way, such as in a tree or a graph, in order to facilitate its processing by specific NLP tasks such as text understanding ones~\citep{RST}. RST, more specifically, is a proposed implementation of this discipline.}, and find that discourse knowledge is captured in intermediate layers, with BERT-based models showing the best results. They even find that static embeddings show a certain degree of rhetorical information encoded within.

\end{itemize}

\subsection{Others}
\label{others}

Some works in our study do not address any specific linguistic level or linguistic phenomenon in particular, but still report interesting findings and reflections, which we summarize in the reminder of this section.

\subsubsection{Knowledge-domain neurons}

\citet{292e02b264e2b608b7f94ceb4e3ba02c8c39420609d170ccc537f1e2f7bbc4b5}, while distancing itself initially from any linguistics-informed approach to interpretability --done so in order not to bias their investigations into these models--, make an interesting discovery regarding specific neurons in PLMs: there seem to be domain-specific or knowledge neurons that are specialized in processing different types of text. For instance, in BERT, their studied model, they found a so-called `social media' neuron which roughly encompasses informal language, a `science' neuron that does the same but for formal and uncommon words, and a `noun' and a `verb' neuron that were highly related with sentences containing large numbers of nouns and verbs respectively. This work even found some neurons as specific as a `United States' one (where words are referred to the topic of the US) or an `Olympic' neuron (where words are related to the Olympics domain). The names and supposed content of these neurons, however, is given by the authors in a qualitative manner, based on a manual inspection of some of the processed examples. The method for discovering these neurons is a marked deviation as well from other works in our survey, as it consists on running a set of sentences through each individual neuron in a model and then clustering them with text mining techniques. This work, other than demonstrating the presupposed existence of these neurons, goes a step further and attempts to perform an \textit{a posteriori} analysis of linguistic knowledge in these models, specifically Part-of-Speech (PoS) tags and sentiment polarity, showing that several of their detected knowledge neurons may be working in combination with others in the same layers to gear the model towards processing these phenomena.

\subsubsection{Multilinguality and the hypothesis of a shared linguistic space in multilingual models}
\label{multilingual_models_shared_space}

A recurring hypothesis regarding linguistic knowledge in multilingual models is that these PLMs encode linguistic structures from different languages in shared subspaces or reusing common internal parameters --either because of efficiency or the presence of common patterns across languages. This is a hypothesis that goes beyond the scope of our survey, but we can briefly refer to some general conclusions on the topic:

\begin{itemize}

\item The Transformer architecture is found to achieve multilinguality thanks to the use of a limited number of parameters, which force models to internally align common structures from different languages~\citep{c150a691ac9f33ab1c91126b1f53eabc2ce350893235c59b1c34187fdf2d6382}.

\item Word order has also been found to be a key aspect for the encoding of multilingual content~\citep{c150a691ac9f33ab1c91126b1f53eabc2ce350893235c59b1c34187fdf2d6382, additional_58}.\footnote{In the same line, an interesting perspective is provided by~\citet{6a9679421f4cc72dbb987a7acbc37cf4863a5dec5d1547db016b8f6215bc8cfc}, which despite their use of a limited methodology --consisting in probing only the 7th layer of an mBERT model--, claim that this model was able to recognize that a series of Slavic languages belong to the same family, but did so as well for German, which despite being Germanic shares a few commonalities with Slavic-type languages --specifically, word order.}

\item Syntactic information is shared across several languages in mBERT~\citep{6c0bcfd8ee5a86207dcb0ac639539437b705ee2b36ce47cc5e645d9d6a521200}.
\item There are reports of a shared syntactic subspace in mBERT, with layers 7 and 8 showing the best results~\citep{additional_58}.

\item In mBERT morphosyntactic information seems to be encoded across languages by the same sets of neurons~\citep{7bca6b86363d9c8ab41964b3f2c90f9e0684e1e0c96e14e3617cefc6ebbf315f} --developing a ``cross-lingually entangled notion of morphosyntax''~\citep{7bca6b86363d9c8ab41964b3f2c90f9e0684e1e0c96e14e3617cefc6ebbf315f}.

\item These shared morphosyntactic encodings are reported as well for the specific phenomena of `subjecthood' (the notion of subject in sentences) and `objecthood'\footnote{\citet{f284d021afd9d17abe71095a7c2879adaf732a37b906048c9796d94deb4eed23} claims these features are represented in a generalized and global way for all the different languages supported by the model, but they are also encoded in a `language-specific enough that it learns language-specific abstract grammatical features'.} (the same, but for objects)~\citep{f284d021afd9d17abe71095a7c2879adaf732a37b906048c9796d94deb4eed23} and agreement~\citep{3d84df6267a2e73cd70576d10a8b6855250044ab8eea8f382060faeec4a09713}.

\end{itemize}

Other works are not as supportive of a supposed multilingual shared space in this type of PLMs, providing some counterarguments:

\begin{itemize}
\item For syntactic agreement, \citet{additional_171} find that although there are sets of neurons commonly associated with this phenomenon across languages in autoregressive multilingual language models --i.e. decoder-based--, they do not find such for masked language models --i.e. encoder-based, such as BERT.

\item \citet{8a88606731bd34583429af7deac06aa5de993adb02d231fad354ac7ad58e91dc} find that multilingual models that had been pre-trained with a clear multilingual objective --LASER (biLSTM-based) and XLM-- tended to encode typological information in the lower layers --and later lost it in higher layers--, but that for models that were trained with a monolingual pre-training objective instead --mBERT and XLM-R-- the results were found to be ``somewhat inconclusive'', with typological information either being captured in lower layers and later transmitted to higher layers, or being evenly spread across the model. This same work, however, also performed several cross-neutralizing experiments that proved that typological information seems to be encoded similarly across languages in all their studied models.

\item \citet{41128f3f273ba991eaff22672cb2714ea9e471d273d3c6f4dcfa590d5e173fdc} find that mBERT, XLM-R and XLM capture well a series of morphological, lexical, word order and syntactic typological features --which are defined in correspondence to those present in two typological datasets: WALS~\citep{WALS} and SSWL (\url{https://terraling.com/groups/7})--, but not so for nominal and verbal features.

Additionally, this same study also reports that their different studied models encode languages and typological features differently, and that the layerwise performance of XLM-R and XLM is stable across languages and their analyzed typological features but is more inconsistent in mBERT.

\item \citet{additional_66}, while mainly using mBERT to --successfully-- deduce a phylogenetic tree for a set of 100 languages, also found that the model performs poorly in the task of cross-lingual semantic retrieval.

\item \citet{additional_61} 
find that mBERT provides worse lexical representations compared to monolingual models.

\item Finally, \citet{ff69f49e3a5e82a147575bc0dfcc94bcf3848fc9c90595829952ec3bbb706041} study polysemy in PLMs and report that this phenomenon seems to not be as well represented in mBERT compared to monolingual BERT-based models. 

\end{itemize}

\section{Discussion}
\label{discussion}

Overall, we have been able to demonstrate a presupposed successful and generalized encoding of many instances of linguistic information of different type --e.g. syntactic, morphological, etc.-- in several Transformer-based models, across different levels: whether in specific neurons, attention heads or layers. However, on a general basis, we have also observed other evidences of general fails in specific aspects of linguistic comprehension in this architecture, ones which would not be normally observed in human speakers. For instance, for syntactic information, there are works that show PLMs displaying agreement errors in some low-frequency verb forms or that these models may be too highly dependent in word order under some circumstances. This evidence could be taken as proof of the generalization of linguistic information in PLMs simply being based off more on statistical relationships rather than actual human-like language rules. This is expected, of course, given the fact that the Transformer is a neural network, and thus is a type of statistical model. Despite this, we must also stress the fact that the studies in this survey, while relying on human theories of language for interpreting the behavior of PLMs, have been able to demonstrate many instances where these models have been able to generalize successfully to these rules. This does not automatically entail, however, an automatic correlation between human knowledge of language and that of PLMs.

We believe, however, that it is possible to reconcile the statistically-based nature of Transformer models with human linguistic theories. This could be done by acknowledging that these models may have come to learn their internal rules of language on their own terms, with those rules being specific to them and not necessarily close to ours. These rules are the interpretation of the patterns that Transformer-based models have been able to deduce to exist in natural language from their seen training texts, which are diverse enough\footnote{This may also explain the need for these models to manage large datasets during their training, as these models must deduce these rules in a generalized and accurate way without any external linguistic cue. As such, low-frequency forms --further hindered by the reliance of models on sparse tokenization schemes such as BPE, which duplicate potential representations of different word forms and/or morphemes-- are likely to produce some generalization errors in these models.} to lead them to learn statistical relationships between tokens which humans can then liken to syntax, semantics, morphology, etc. Although these models do not understand per se these high-level linguistic concepts\footnote{As neither do humans without a background on formal linguistics anyway.}, and that they encode linguistic information of different kind and levels using the same self-attention mechanism --which reduces all this information to the same level--, the generalized observations seem to be in most cases indicative of the learned statistical relationships to be similar to some theories of language. It is, thus, a PLM's interpretation of the underlying rules of human language, including potential generalization errors, and it is likely in this sense that more pre-training texts could help further reduce these types of errors.

An issue present across many of the studies we analyzed is that the detection of a specific language phenomenon in a model's parameters does not necessarily imply that the model might be using that information during inference. This is commonly presented as a dichotomy between `correlation' and `causation' in these models' representations~\citep{belinkov-2022-probing}. 
This has led to the creation, in the area of probes, of a new type called `causal probes'~\citep{additional_161, 0a8f4755cc0cdbfa5ba0f95dfda7061bd32f36a8c81e586fffed249ec106da83}, that attempt to demonstrate whether that information is truly used by a model. Some examples of causal probes in use by papers in our survey include amnesic probes~\citep{de14b3483b08a04f231f0908122bf595363d9bf8fb31c2840302d2d56a3be625, additional_178}, information-theoretic probes~\citep{additional_54} or dropout probes~\citep{additional_157}.
Another issue, exposed by~\citet{probesnotgeneralize}, is that the representations learned by small models of this type --not only probes-- might not be truly indicative on how a bigger, analyzed model might truly generalize to a specific type of learned information, since a proposed generalization as obtained from these classifiers might not apply to out-of-distribution samples. Similarly, \citet{additional_114} also report that probes may simply be learning all information required for successful probing results from the linear context surrounding the tokens, without implicitly learning any sort of structured linguistic information. Other work also point to a related pitfall across many probing works in that they only rely on using Accuracy in their probing results as their single metric to measure the linguistic competence of a source PLM~\citep{hewitt-liang-control-probes, additional_80, additional_54}.

When interpreting a pretrained model, we need to consider that it will have to be eventually adapted to a downstream task. It is not unlikely to assume that an observed linguistic pattern in a pretrained model might be `erased' when the model is further trained. Many studies, in this sense, attempt to analyze PLMs both in pre-trained and fine-tuned settings --e.g.~\citep{additional_150, 4a343f0a56c9dea29ff8484c457c3a894f5b8d37c85f407b2f7723c96ccc3494, additional_64}. In this line, \citet{additional_157} discovers that some fine-tuned models still preserve and use information stemming from pre-trained models, while~\citet{additional_64} report the model losing part of this information when fine-tuned on a downstream NLI task. This research question, however, is outside the scope of this survey and is left to be covered by other works.

\section{Related work}
\label{related_work}

Our analysis is close in its aims to the research work presented by~\citet{belinkov-glass-2019-analysis}, which also analyzes linguistic knowledge in artificial neural networks and presents some common methodologies with our work. Their research, however, involves exclusively pre-Transformer architectures, such as Long-Short Term Memory (LSTM) models, or dedicated Machine Translation systems --e.g.~\citep{Bahdanau-attention}.
Our work also builds upon the survey presented by~\citet{bertology} --the work that introduced the field of \textit{BERTology}--, sharing with it many of its objectives and methodologies. Our proposal provides an up-to-date revision to their overall conclusions with newer works, PLM architectures and methodologies, as well as additional, more in-depth information on the linguistic capabilities of these models. Our survey includes 141 articles not covered by~\citet{bertology}.

A recent, similar work to our proposal is that of~\citet{XAILING2024}, which also studies the linguistic competences of PLMs across multiple levels --e.g. syntax, semantics, etc. That survey, as in our work, studies pre-trained language models without any sort of post-hoc modifications --e.g. fine-tuning. Their work, however, involves English-based models exclusively and, most importantly, only focuses on behavioral studies of PLMs, where these models are analyzed in a black-box setting. Our work, on the other hand, attempts to go beyond this paradigm and presents research that analyzes PLMs using internal representations. 

Another similar approach is that of~\citet{XAILING2024_2_HOLMES}, which also aims to study linguistic competence in PLMs and present a unified framework for that purpose. Similar to our work, they also present a survey of existing works in the area and focus on those that provide explanations that explore models' internal representations. While coinciding in many research points and objectives with our proposal, that work limits itself to the analysis of probing-based studies only and does not explore models outside of those trained for the English language. Additionally, they also fail to deliver a unified account of the general linguistic knowledge seemingly present in PLMs, not providing an overall reflection on the general linguistic capabilities of the Transformer architecture. The main purpose of that research work is to present a linguistic interpretability framework that attempts to unify ongoing research in the area.

\citet{XAILING2023} also presents some research works into linguistic interpretability, but does so in a limited manner, with a fewer number of research studies mentioned, and encompassing examples of both behavior-based studies as well as those that rely on internal representations.

\section{Conclusions}
\label{conclusions}

In this survey we have presented a series of research work that attempt to discover how linguistic knowledge may be present inside modern, state-of-the-art PLMs based on the Transformer architecture. We have presented their overall conclusions in an organized manner, from the perspective of different traditional linguistic disciplines such as syntax or lexico-semantics, across several languages with different typologies. Overall, we have been able to report many instances of well-encoded linguistic phenomena inside these models, but also of systematic fails of some other phenomena at the same time. The provided conclusions on linguistic competence of these models, although contradicting at times, seem to point nevertheless in the direction of an architecture which is able to generalize certain aspects of human language rules, roughly corresponding to the same ones seen on traditional linguistics, which it seems to deduce simply from the pre-training texts --of different quality-- it has seen. It achieves this linguistic proficiency, however, in its own terms --i.e. via statistically-based methods-- and with some errors along the way.

It is unclear as which future tendencies there will be in the study of linguistic interpretability in PLMs. The rise of increasingly larger autoregressive models may mean the appeareance of more prompting-like studies in the near future, whereas probing and related techniques might become increasingly phased out due to the amount of computation needed to deploy them in LLMs~\citep{Explainability_for_Large_Language_Models_A_Survey}. This is already happening in some degree in our observed list of papers, as many of the analysis we have found are done in BERT-based models and omit more modern ones. 
Perhaps ever-larger PLMs may become increasingly unaffordable to pre-train at some point and the industry (particularly small and medium-sized enterprises) may shift to smaller-sized models. This could render our analyzed interpretability techniques still suitable in this hypothetical scenario.

\bibliographystyle{ACM-Reference-Format}
\bibliography{sample-base}

\end{document}